\documentclass[letterpaper, 10 pt, conference]{ieeeconf}    

\IEEEoverridecommandlockouts
     
\usepackage{amsmath,amssymb,mathtools}
\usepackage{bm}
\usepackage{tabularx}
\usepackage{booktabs}
\usepackage{xcolor}
\usepackage{soul}

\usepackage{graphicx} 
\usepackage{caption}
\usepackage{subcaption}
\usepackage{subcaption}
\let\labelindent\relax
\usepackage{enumitem}
\usepackage{hyperref}
\hypersetup{
    pdfauthor={},
    pdftitle={},
    pdfsubject={},
    pdfkeywords={},
    colorlinks=false, 
    pdfborder={0 0 0}
}

\usepackage{algorithm}
\usepackage{algpseudocode}

\usepackage[
    style=ieee, 
    backend=biber, 
    natbib=true,
    doi=false,   
    url=false,   
    isbn=false,  
    eprint=false 
]{biblatex}

\DeclarePairedDelimiter{\norm}{\lVert}{\rVert}
\newcommand{\atanTwo}{\operatorname{atan2}}
\newcommand{\SO}{\mathrm{SO}(3)}

\newcommand{\mytitle}{\LARGE \bf Geometric Inverse Flight Dynamics on \(\SO\) and Application to Tethered Fixed-Wing Aircraft}
\title{\mytitle}

\author{%
Antonio Franchi$^{1,2}$ and Chiara Gabellieri$^{1}$%
\thanks{$^{1}$ Robotics and Mechatronics lab, Faculty of Electrical Engineering, Mathematics \& Computer Science, University of Twente, Enschede, The Netherlands. \href{mailto:schol@r-franchi.eu}{schol@r-franchi.eu}, \href{mailto:c.gabellieri@utwente.nl}{c.gabellieri@utwente.nl}}%
\thanks{$^{2}$Department of Computer, Control and Management Engineering, Sapienza University of Rome, 00185 Rome, Italy.}%
}

\newif\ifarxiv
\arxivtrue  

\begin{document}
\maketitle
\begin{refsection}

\begin{abstract}
We present a robotics-oriented, coordinate-free formulation of inverse flight dynamics for fixed-wing aircraft on $\mathrm{SO}(3)$. Translational force balance is written in the world frame and rotational dynamics in the body frame; aerodynamic directions (drag, lift, side) are defined geometrically, avoiding local attitude coordinates. Enforcing coordinated flight (no sideslip), we derive a closed-form trajectory-to-input map yielding the attitude, angular velocity, and thrust–angle-of-attack pair, and we recover the aerodynamic moment coefficients component-wise. Applying such a map to tethered flight on spherical parallels, we obtain analytic expressions for the required bank angle and identify a specific \emph{zero-bank locus} 
where the tether tension exactly balances centrifugal effects, highlighting the decoupling between aerodynamic coordination and the apparent gravity vector. Under a simple lift/drag law, the minimal-thrust angle of attack admits a closed form. These \emph{pointwise quasi‑steady inversion} solutions become steady-flight trim when the trajectory and rotational dynamics are time-invariant. The framework bridges inverse simulation in aeronautics with geometric modeling in robotics, providing a rigorous building block for trajectory design and feasibility checks.
\end{abstract}

\begin{center}
\emph{Supplementary Material available at\\ \url{https://arxiv.org/abs/2602.17166}}
\end{center}

\section{Introduction}

Aerial manipulation with multirotors has seen rapid progress, but endurance constraints limit long-duration tasks.
To overcome this, recent works introduced the \emph{non-stop flight} paradigm, where multiple carriers move perpetually while holding a cable-suspended load at a fixed pose~\cite{GabellieriFranchiICUAS2024,GabellieriFranchiArxiv2025}. 
Those results show that, for three or more carriers, one can synthesize coordinated, phase-shifted, cyclic trajectories that keep the load static while each carrier never stops, thus enabling energy-efficient manipulation beyond multirotor hover endurance. 
However, in that line of work, the dynamics of each carrier are \emph{abstracted as double integrators}, i.e., the aerodynamic and rigid-body actuation dynamics of aircraft are not included.

Robotics formulations for \emph{quadrotor} aerial manipulation of cable-suspended loads model a thrust-vectoring rotorcraft on \( \mathrm{SE}(3) \) coupled to a pendular load on \( \mathrm{S}^2 \), 
enabling differential-flatness and geometric-control design~\cite{SreenathMichaelKumarICRA2013,li2021cooperative,sun2025agile}.
While powerful for multirotors, these models \emph{do not} capture fixed-wing aerodynamics (lift, angle of attack, aerodynamic moment coupling, coordinated-flight constraint), and are therefore \emph{not directly applicable} to fixed-wing carriers. 
This motivates a geometric, world-frame composition with explicit aerodynamic directions suitable for fixed-wing inverse synthesis.

\emph{Related Work.}
To the best of our knowledge, \emph{aircraft-based manipulation} remains essentially unexplored within the robotics community. Arguably, the only preliminary attempt is~\cite{williams2009dynamics}, which, however, neglects the dynamic coupling between the cable and the aircraft.
By contrast, the airborne wind energy (AWE) literature has long studied tethered wings and their flight mechanics~\cite{FagianoTCST2014}, but not the robotics-oriented inverse synthesis that produces realizable input and orientation trajectories directly from task-space specifications (and certainly not for manipulation).
At the same time, the aeronautics community offers (i) \emph{inverse simulation} methods that recover control histories using local angles~\cite{BlajerInverseSim2009,MarzoukInverseSim2024}; (ii) \emph{nonlinear dynamic inversion} (NDI) and incremental NDI (INDI) for control-law design~\cite{MillerNDI2011,smeur2016adaptive,SteensenINDI2022}; and (iii) coordinated-turn relations linking curvature, load factor, and bank~\cite{StengelFlightDynamics2022,StevensLewisJohnson2015}.
Yet, a concise \emph{robotics-oriented, coordinate-free, \( \SO \)-based} inverse synthesis for fixed-wing aircraft is still missing.

Relative to classical inverse simulation~\cite{BlajerInverseSim2009,MarzoukInverseSim2024}, our approach is \emph{coordinate-free} and world-frame geometric, making it directly accessible for robotics planning (as in multirotor flatness methods~\cite{MellingerICRA2011,FaesslerFranchiScaramuzza2018,LeeLeokMcClamrochCDC2010}).
Relative to NDI/INDI~\cite{MillerNDI2011,smeur2016adaptive}, we focus on \emph{trajectory inversion} and synthesis of the attitude, angular velocity, thrust, and angle of attack 
 rather than a control-law structure.
Relative to AWE~\cite{FagianoTCST2014}, we target robotics use-cases that require geometric inverse synthesis with explicit feasibility checks and manipulation-oriented external forces. Differently from control-line aero models, where the pilot controls the aerodynamic surfaces of the aircraft using one or more tethers \cite{zuhdi2023ratio}, in our model, the tether is attached at the  center of mass, and it is not directly connected to the inputs of the aircraft. 
Such a modeling choice is instrumental for the manipulation of cable-supended loads. 
Crucially, this work complements our prior non-stop flight results~\cite{GabellieriFranchiICUAS2024,GabellieriFranchiArxiv2025} by contributing the aircraft-level building block needed to move from abstractions toward \emph{aircraft-based} manipulation.

The main contributions of the work are as follows.
\emph{(1) Geometric \( \SO \) model for fixed-wing aircraft.}
We formulate Newton–Euler dynamics in a \(z\)-up world frame, with globally valid aerodynamic directions and rate-damping moments, remaining compatible with standard coefficient maps and stability derivatives~\cite{StengelFlightDynamics2022,EtkinReid1995}. 
 We believe that such a problem formulation may represent a useful starting point for the robotics community to design motion and interaction controllers.
\emph{(2) Inverse flight dynamics (IFD) in closed form.}
Given a desired Center of Mass (CoM) trajectory and external loads, we construct the attitude 
 and body angular velocity 
  geometrically, and solve the 2D force balance for 
  the thrust and the angle of attack under coordinated flight.
We also recover the required aerodynamic moment coefficients 
 component-wise.
\emph{(3) Tethered fixed-wing on a spherical parallel.}
We apply the theory to an aircraft tethered to the origin, flying on a spherical parallel at constant speed, and subject to the cable tension.
We derive analytic expressions for the required bank angle, identifying a specific \emph{zero-bank locus} where the tether tension exactly counteracts centrifugal effects, decoupling aerodynamic coordination from the apparent gravity vector.
\emph{(4) Minimal-thrust angle of attack.}
Under a simple lift–drag law, the 
 value of the angle of attack that minimizes thrust admits a closed-form solution (Cardano), yielding constant angle of attack and minimum thrust 
  for constant-speed circular motion.


\section{Geometric Newton--Euler Model on \(\SO\)}
\label{sec:model}

All translational force balances are written in the \emph{world} frame (gravity, aerodynamics, propulsion, external), with body-frame contributions mapped via the aircraft rotation matrix. Rotational dynamics are expressed in the \emph{body} frame using Euler’s equations. Aerodynamic directions (drag, lift, side) are defined purely from vectors (flow, chord, span, wing normal), yielding a globally valid model suitable for geometric control, estimation, and robotics applications.


In the following, boldface letters denote vectors and superscripts \((\cdot^W,\cdot^B)\) the frame where a quantity is expressed. For any \(\mathbf{x}\in\mathbb{R}^3\), \([\mathbf{x}]_\times\) is the skew operator, with \([\mathbf{x}]_\times \mathbf{y} = \mathbf{x}\times \mathbf{y}\).
We define the following reference frames:
\begin{itemize}[leftmargin=*]
    \item World frame \(W\): orthonormal basis \(\{\mathbf{e}_1^W,\mathbf{e}_2^W,\mathbf{e}_3^W\}\) with \(\mathbf{e}_3^W\) pointing up; gravity \(\mathbf{g}^W = -g\,\mathbf{e}_3^W\).
    \item Body frame \(B\): attached to the aircraft with origin in the CoM, and canonical axes
    \(\mathbf{e}_1^B\) (forward/chord),
    \(\mathbf{e}_2^B\) (span/left),
    \(\mathbf{e}_3^B\) (wing-plane normal/up).
\end{itemize}
The rotation matrix \(R\in \SO\) maps body to world:
if \(\mathbf{x}^B\) is expressed in frame \(B\), its world expression is \(\mathbf{x}^W = R\,\mathbf{x}^B\).

\subsubsection*{Forward-Left-Up (FLU) vs.\ aeronautics Forward-Right-Down (FRD)}
We adopt the robotics-oriented FLU convention, whereas many aeronautics texts use FRD. This implies a sign flip in the \(y\)-axis relative to standard flight-mechanics moment conventions (e.g., the sign of \(C_m\) when mapped to pitch). 
When comparing with aeronautics databases, care must be taken to account for this axis orientation difference.

\ifarxiv
For the reader's convenience, a summary of the nomenclature and the constitutive model equations (derived hereafter) is provided in Tables~\ref{tab:symbols} and~\ref{tab:model_summary} of the {supplementary material at the end of this document}.
\else{For the reader's convenience, a summary of the nomenclature and the constitutive model equations (derived hereafter) is provided in Tables II and III of the supplementary material available at \url{http://arxiv.org/abs/2602.17166}.}
\fi


We work directly on \(\SO\) (no local attitude coordinates):
\begin{equation}
\dot{\mathbf{p}}^W = \mathbf{v}^W,\qquad
\dot{R} = R\,[\boldsymbol{\omega}^B]_\times.
\label{eq:kinematics}
\end{equation}

The translational dynamics in the world frame is:
\begin{equation}
m\,\dot{\mathbf{v}}^W
= m\,\mathbf{g}^W
\;+\; R\!\left(\mathbf{F}_{\text{prop}}^B + \mathbf{F}_{\text{aero}}^B\right)
\;+\; \mathbf{f}_{\text{ext}}^W
\label{eq:translation}
\end{equation}
and the rotational dynamics in the body frame is:
\begin{equation}
I_B\,\dot{\boldsymbol{\omega}}^B
\;+\;\boldsymbol{\omega}^B\times\!\left(I_B\,\boldsymbol{\omega}^B\right)
= \boldsymbol{\tau}_{\text{prop}}^B
\;+\;\boldsymbol{\tau}_{\text{aero}}^B
\;+\;\boldsymbol{\tau}_{\text{ext}}^B.
\label{eq:rotation}
\end{equation}
Since \(\mathbf{f}_{\text{ext}}^W\) is applied at the CoM, it produces no moment; if a known force is applied at a point \(r^B\) (expressed in \(B\)), its equivalent at the CoM is \(\mathbf{f}_{\text{ext}}^W\) plus the moment \(r^B\times R^\top\mathbf{f}_{\text{ext}}^W\) added to \(\boldsymbol{\tau}_{\text{ext}}^B\).

Define the air-relative velocity vectors in $W$ and $B$ as:
\begin{equation}
\mathbf{v}_a^W \!=\! \mathbf{v}^W - \mathbf{w}^W,\quad
\mathbf{e}_a^W \!=\! \mathbf{v}_a^W / \|\mathbf{v}_a^W\|\ \ (\|\mathbf{v}_a^W\|>0),
\label{eq:airrel_W}
\end{equation}
\begin{equation}
\mathbf{v}_a^B \!=\! R^\top \mathbf{v}_a^W,\quad
\mathbf{e}_a^B \!=\! R^\top \mathbf{e}_a^W\ \ (\|\mathbf{v}_a^B\|>0),
\label{eq:airrel_B}
\end{equation}
with $\mathbf{w}^W$ being the wind velocity, and the dynamic pressure
\begin{equation}
q = \tfrac{1}{2}\,\rho\,\|\mathbf{v}_a^B\|^2.
\label{eq:q}
\end{equation}
 When \(\|\mathbf{v}_a^B\|\) is small, define \(\|\mathbf{v}_a^B\|_\epsilon := \max(\|\mathbf{v}_a^B\|,\epsilon)\) and use
\(\tilde{\mathbf{e}}_a^B := \mathbf{v}_a^B / \|\mathbf{v}_a^B\|_\epsilon\), \( \tilde{q} := \tfrac12 \rho \|\mathbf{v}_a^B\|_\epsilon^2\), with a small \(\epsilon>0\) to avoid ill-conditioning.

Let \(\mathbf{c}^B\) (chord/forward), \(\mathbf{s}^B\) (span/left), and \(\mathbf{n}^B\) (wing-plane normal/up) be a right-handed (in such order) triad of unit vectors fixed in \(B\). We have \((\mathbf{c}^B,\mathbf{s}^B,\mathbf{n}^B)=(\mathbf{e}_1^B,\mathbf{e}_2^B,\mathbf{e}_3^B)\), corresponding to a  FLU convention.

Define the aerodynamic directions in \(B\) as:
\begin{equation}
\mathbf{e}_D^B := -\,\mathbf{e}_a^B; \, 
\mathbf{e}_L^B := \tfrac{\mathbf{n}^B - (\mathbf{n}^B\!\cdot\!\mathbf{e}_a^B)\,\mathbf{e}_a^B}
                        {\big\|\mathbf{n}^B - (\mathbf{n}^B\!\cdot\!\mathbf{e}_a^B)\,\mathbf{e}_a^B\big\|};\,
\mathbf{e}_Y^B :=\mathbf{e}_a^B \times \mathbf{e}_L^B. \label{eq:dragliftsidedir}
\end{equation}
The vectors \((\mathbf{e}_D^B,\mathbf{e}_Y^B,\mathbf{e}_L^B)\)  form a right-handed orthonormal set aligned with drag, side-force, and lift directions. When using the regularized \(\tilde{\mathbf{e}}_a^B\), replace \(\mathbf{e}_a^B\) by \(\tilde{\mathbf{e}}_a^B\) in \eqref{eq:dragliftsidedir}.

\subsubsection*{Remark (frame origin).}
We do not assign an “aerodynamic frame origin”: all net aerodynamic forces are applied at the CoM in \eqref{eq:translation}, and all net aerodynamic moments about the CoM are modeled in \eqref{eq:TaeroB}. If distributed lifting surfaces are modeled, moments arise from \(r_i^B\times \mathbf{F}_i^B\) (see below).

\subsubsection*{Remark (Lift definition).}
Unlike standard conventions where Lift is typically defined as the component of aerodynamic force perpendicular to the relative wind and is constrained to the vehicle’s symmetry plane, \eqref{eq:dragliftsidedir} ensures \(\mathbf{e}_L^B\) is strictly orthogonal to the velocity vector \(\mathbf{e}_a^B\). We adopt a global geometric basis \((\mathbf{e}_D^B,\mathbf{e}_Y^B,\mathbf{e}_L^B)\) that is singularity-free and coordinate-free (preferred in robotics); under coordinated flight (\(\beta=0\)) the two conventions coincide, while for \(\beta\neq 0\) the decomposition differs by the symmetry-plane constraint.

\subsubsection*{Angle of Attack (AoA)}

Define the angle of attack as
\begin{equation}
\alpha \;=\; \atanTwo\!\Big(-\,\mathbf{e}_a^B\!\cdot\!\mathbf{n}^B,\ \mathbf{e}_a^B\!\cdot\!\mathbf{c}^B\Big),
\label{eq:alpha}
\end{equation}
so that positive \(\alpha\) corresponds to a flow component with negative projection on \(\mathbf{n}^B\) and positive on \(\mathbf{c}^B\) (nose-up relative to flow), matching the standard aerodynamic interpretation.

\subsubsection*{Sideslip Angle} Define the sideslip angle
\begin{equation}
\beta \;=\; \atanTwo\!\Big(-\,\mathbf{e}_a^B\!\cdot\!\mathbf{s}^B,\ \sqrt{\,1-(\mathbf{e}_a^B\!\cdot\!\mathbf{s}^B)^2\,}\Big),
\label{eq:beta}
\end{equation}
i.e., the signed angle between \(\mathbf{e}_a^B\) and the sagittal plane spanned by \((\mathbf{c}^B,\mathbf{n}^B)\). Under coordinated flight, \(\beta=0\).
Note the negative sign: since \(\mathbf{s}^B\) points Left, a positive projection \(\mathbf{e}_a^B\!\cdot\!\mathbf{s}^B\) implies motion to the left (wind from port). The negative sign ensures that \(\beta > 0\) corresponds to motion to the right (wind from starboard), as in standard aeronautical conventions.

\subsubsection*{Aerodynamic Forces and Moments}

Compute forces in \(B\) using the geometric directions:
\begin{align}
\mathbf{F}_{\text{aero}}^B
&= q\,S\Big(
C_D(\xi)\,\mathbf{e}_D^B
\;+\;
C_L(\xi)\,\mathbf{e}_L^B
\;+\;
C_Y(\xi)\,\mathbf{e}_Y^B
\Big),
\label{eq:FaeroB}
\end{align}
where $C_D$, $C_L$, and $C_Y$ are all functions of $\xi:=(\alpha,\beta,\mathbf{u})$
with \(\mathbf{u}\) being the control inputs  (elevator, aileron, rudder, flaps, etc.). Coefficients may depend on \(\alpha,\beta\), Mach, Reynolds, and control deflections. Map to world via
\begin{equation}
\mathbf{F}_{\text{aero}}^W = R\,\mathbf{F}_{\text{aero}}^B.
\label{eq:FaeroW}
\end{equation}

The net aerodynamic moments about the CoM are
\begin{equation}
\boldsymbol{\tau}_{\text{aero}}^B
= q\,S\,\big(
C_\ell\,b\,\mathbf{e}_1^B
\;+\;
C_m\,\bar{c}\,\mathbf{e}_2^B
\;+\;
C_n\,b\,\mathbf{e}_3^B
\big)
\;+\; D_\omega\,\boldsymbol{\omega}^B,
\label{eq:TaeroB}
\end{equation}
where \(b\) is the span, \(\bar c\) the mean aerodynamic chord, \(C_\ell,C_m,C_n\) the roll/pitch/yaw moment coefficients (also functions of $\xi$), and \(D_\omega\preceq 0\) a rate-damping matrix capturing stability derivatives (e.g., \(l_p,m_q,n_r\)).%
\footnote{If preferred, one can adopt a single reference length \(\bar c\) for all three components, at the cost of reinterpreting the numerical coefficients.}

\subsubsection*{Distributed alternative.}
For multiple lifting/dragging surfaces indexed by \(i\), with positions \(r_i^B\) from the CoM, use the apparent local air velocity
$
\mathbf{v}_{a,i}^B = \mathbf{v}_a^B + \boldsymbol{\omega}^B \times r_i^B,
$
construct \(\mathbf{F}_i^B\) via \eqref{eq:FaeroB} using \(\mathbf{v}_{a,i}^B\), and sum
\(\boldsymbol{\tau}_{\text{aero}}^B=\sum_i r_i^B \times \mathbf{F}_i^B\).

\subsubsection*{Propulsion (thrust)}
Let the thrust direction be \(\mathbf{u}_t^B\) (unit, e.g., \(\mathbf{u}_t^B\approx+\mathbf{e}_1^B\) for a tractor propeller). Then
\begin{equation}
\mathbf{F}_{\text{prop}}^B = T\,\mathbf{u}_t^B,\qquad
\boldsymbol{\tau}_{\text{prop}}^B = \boldsymbol{\tau}_{\text{prop}}^B(\mathbf{u}),
\label{eq:prop}
\end{equation}
with \(T\) and \(\boldsymbol{\tau}_{\text{prop}}^B\) given by the propulsion system and inputs. Propwash can be modeled by augmenting \(\mathbf{v}_a^B\) near the propeller or absorbed into the coefficient maps.

\subsubsection*{External force and moment}
A known external force at the CoM, expressed in the world frame, enters \eqref{eq:translation} as \(\mathbf{f}_{\text{ext}}^W\). A known external moment at the CoM, expressed in the body frame, enters \eqref{eq:rotation} as \(\boldsymbol{\tau}_{\text{ext}}^B\). If the line of action does not pass through the CoM, model the equivalent moment as previously described.

\section{Geometric Inverse Flight Dynamics}
\label{sec:inverse}

This problem is known in aeronautics as \emph{inverse simulation}. 



\textbf{Given:}
\begin{itemize}
  \item The CoM trajectory \(\mathbf{p}^W(t)\) with \(\mathbf{v}^W(t)=\dot{\mathbf{p}}^W(t)\), \(\mathbf{a}^W(t)=\dot{\mathbf{v}}^W(t)\); higher derivatives available if needed.
  \item External loads at the CoM: \(\mathbf{f}_{\text{ext}}^W(t)\in\mathbb{R}^3\), \(\boldsymbol{\tau}_{\text{ext}}^B(t)\in\mathbb{R}^3\).
  \item Wind field \(\mathbf{w}^W(t)\) (possibly time-varying; if unknown, set \(\mathbf{w}^W=\mathbf{0}\)).
  \item Aircraft parameters: \(m\), \(I_B\succ 0\), aerodynamic references \((\rho,S,\bar c,b)\), and aerodynamic rate damping \(D_\omega\succeq 0\).
  \item Thrust direction equals the body chord: \(\mathbf{u}_t^B=\mathbf{c}^B\) (tractor configuration), hence \(\mathbf{u}_t^W=R\,\mathbf{c}^B=\mathbf{c}^W\).
\end{itemize}

\textbf{Find:}
\begin{enumerate}
  \item Orientation \(R(t)\in\SO\) and angular velocity \(\boldsymbol{\omega}^B(t)\).
  \item Control histories \(T(t)\ge 0\), \(C_D(t),C_L(t),C_Y(t)\), \(C_\ell(t),C_m(t),C_n(t)\) that satisfy the dynamics.
\end{enumerate}

When feasible, we aim for coordinated flight (\(\beta=0\)).

\subsection{Closed-form construction at each time \(t\)}

\paragraph*{Step 0: Air-relative velocity and dynamic pressure.}
Compute air-relative direction vector $\mathbf{e}_a^W$ from~\eqref{eq:airrel_W} (regularize near zero airspeed if needed),  and dynamic pressure \(q\) from~\eqref{eq:q}.

\subsubsection*{Step 1: Required net force from the trajectory.}
Define the world-frame \emph{required} net force that must be generated by propulsion+aerodynamics:
\begin{equation}
\mathbf{F}_{\text{req}}^W
= m\,\mathbf{a}^W \;-\; m\,\mathbf{g}^W \;-\; \mathbf{f}_{\text{ext}}^W.
\label{eq:F_req_wind}
\end{equation}
(\emph{Note:} wind does not appear in \(\mathbf{F}_{\text{req}}^W\); it only affects the aerodynamic directions and the dynamic pressure.)

\subsubsection*{Step 2: Decompose along the air-relative direction.}
Decompose \(\mathbf{F}_{\text{req}}^W\) into components along perpendicular to \(\mathbf{e}_a^W\):
\begin{equation}
f_\parallel := \mathbf{F}_{\text{req}}^W\!\cdot\!\mathbf{e}_a^W,\;\;
\mathbf{F}_{\perp}^W := \mathbf{F}_{\text{req}}^W - f_\parallel\,\mathbf{e}_a^W,\;\;
f_\perp := \big\|\mathbf{F}_{\perp}^W\big\|.
\label{eq:decompose_wind}
\end{equation}

\subsubsection*{Step 3: Define the trajectory frame and impose coordinated flight.}
If \(f_\perp>0\), we identify the direction of the required lift (trajectory normal) and the binormal. To ensure a right-handed system consistent with a \emph{z-up} robotics convention, we define the binormal \(\mathbf{s}_{\text{traj}}^W\) to complete the FLU triad \(\{\mathbf{e}_a^W, \mathbf{s}_{\text{traj}}^W, \mathbf{n}_{\text{traj}}^W\}\):
\begin{equation}
\mathbf{n}_{\text{traj}}^W := \mathbf{F}_{\perp}^W / \|\mathbf{F}_{\perp}^W\|,\qquad
\mathbf{s}_{\text{traj}}^W := \mathbf{n}_{\text{traj}}^W \times \mathbf{e}_a^W.
\label{eq:traj_frame}
\end{equation}
Note that \(\mathbf{s}_{\text{traj}}^W\) points to the aircraft’s \emph{left} (port side).
Here, \(\mathbf{n}_{\text{traj}}^W\) represents the direction in which the aerodynamic lift (mostly) and thrust (slightly) must act to satisfy the trajectory requirement, and \(\mathbf{s}_{\text{traj}}^W\) is the binormal vector of the curve relative to the air mass.

\subsubsection*{Coordinated Flight.}
We define coordinated flight as \(\beta=0\) \emph{(aerodynamic coordination)}. Geometrically, this requires the body \(y\)-axis to align with the trajectory binormal:
\begin{equation}
\mathbf{s}^W \equiv \mathbf{s}_{\text{traj}}^W \quad \text{and} \quad \mathbf{n}_{\text{curve}}^W \equiv \mathbf{n}_{\text{traj}}^W.
\label{eq:coordinated_imposition}
\end{equation}
This construction ensures that \(\{\mathbf{e}_a^W,\mathbf{s}^W,\mathbf{n}_{\text{curve}}^W\}\) forms an orthonormal, right-handed triad where the required force lies entirely within the \(\mathbf{e}_a^W\)-\(\mathbf{n}_{\text{curve}}^W\) plane, thereby reducing the thrust/AoA determination to a 2D balance.

\subsubsection*{Step 4: Complete the body axes with the to-be-chosen angle of attack \(\alpha\).} 
With \(\beta=0\), the body axes are obtained by rotating the trajectory frame about the \(y\)-axis (\(\mathbf{s}^W\)) by the to-be-chosen angle of attack \(\alpha\). This pitches the nose \(\mathbf{c}^W\) “up” towards \(\mathbf{n}_{\text{curve}}^W\):
\begin{align}
\begin{aligned}
\mathbf{c}^W &= \cos\alpha\,\mathbf{e}_a^W \;+\; \sin\alpha\,\mathbf{n}_{\text{curve}}^W, \\
\mathbf{n}^W &= -\sin\alpha\,\mathbf{e}_a^W \;+\; \cos\alpha\,\mathbf{n}_{\text{curve}}^W.
\end{aligned}
\label{eq:axes_world_wind}
\end{align}
The resulting attitude matrix is constructed as:
\begin{equation}
R(t) = \begin{bmatrix}
\mathbf{c}^W(t) & \mathbf{s}^W(t) & \mathbf{n}^W(t)
\end{bmatrix}\in\SO.
\label{eq:R_closed_wind}
\end{equation}
This construction guarantees \(\det(R)=+1\) and \(\mathbf{c}^W = \mathbf{s}^W \times \mathbf{n}^W\), consistent with the right-hand rule.

\subsubsection*{Step 5: Aerodynamic force approximation.}
To decouple the translational dynamics (slow timescale) from the rotational dynamics (fast timescale), we assume that the control surfaces \(\mathbf{u}\) primarily generate moments and have a negligible effect on the total lift and drag compared to \(\alpha\). 
 Under this approximation and given that \(\beta=0\), we have \(C_L(\alpha,\beta,\mathbf{u}) \approx C_L(\alpha)\) and \(C_D(\alpha,\beta,\mathbf{u}) \approx C_D(\alpha)\)). Thus,  the aerodynamic control input $\mathbf{u}$ disappears from the lift equations and is primarily used to control the aircraft attitude as in~\eqref{eq:u_solver}.

 The aerodynamic force is then:
\begin{equation}
\mathbf{F}_{\text{aero}}^W
= q\,S\left(C_L(\alpha)\,\mathbf{n}_{\text{curve}}^W \;-\; C_D(\alpha)\,\mathbf{e}_a^W\right),
\label{eq:Faero_coord_wind}
\end{equation}
since \(\mathbf{e}_L^W=\mathbf{n}_{\text{curve}}^W\) and \(\mathbf{e}_D^W=-\mathbf{e}_a^W\).

\subsubsection*{Step 6: Thrust decomposition and solution for \(T, \alpha\).}
Thrust acts along \(\mathbf{c}^W\): \(\mathbf{F}_{\text{prop}}^W = T\,\mathbf{c}^W
= T\big(\cos\alpha\,\mathbf{e}_a^W + \sin\alpha\,\mathbf{n}_{\text{curve}}^W\big)\).
Projecting \(\mathbf{F}_{\text{req}}^W=\mathbf{F}_{\text{prop}}^W+\mathbf{F}_{\text{aero}}^W\) on \(\mathbf{e}_a^W\) and \(\mathbf{n}_{\text{curve}}^W\) yields:
\begin{equation}
\begin{aligned}
f_\parallel &= T\cos\alpha \;-\; q\,S\,C_D(\alpha),\, f_\perp \ \ &= T\sin\alpha \;+\; q\,S\,C_L(\alpha).
\end{aligned}
\label{eq:two_eqs_wind}
\end{equation}
Define the required thrust components as:
\begin{equation}
T_\parallel(\alpha) := f_\parallel + q\,S\,C_D(\alpha),\;\;
T_\perp(\alpha) := f_\perp - q\,S\,C_L(\alpha).
\label{eq:T_components_wind}
\end{equation}
\(T\) and \(\alpha\) are found by solving the fixed-point equations:
\begin{equation}
T = \sqrt{T_\parallel(\alpha)^2 + T_\perp(\alpha)^2},\;\;
\alpha = \atanTwo\big(T_\perp(\alpha),\;T_\parallel(\alpha)\big).
\label{eq:T_alpha_closed_wind}
\end{equation}
Since \(C_L, C_D\) depend on \(\alpha\), \eqref{eq:T_alpha_closed_wind} is typically solved via Newton–Raphson iteration or by using the previous time-step’s \(\alpha\) for the coefficients. These are pointwise (quasi-steady) solutions for \((T,\alpha)\) (instantaneous inversion) consistent with aerodynamic coordination \(\beta=0\), not necessarily a steady-flight \emph{trim} in the aeronautic sense, unless the trajectory and rotational dynamics are steady.

\subsubsection*{Step 7: Angular velocity \(\boldsymbol{\omega}^B(t)\).}
The angular velocity associated with \(R(t)\) is:
\begin{equation}
\boldsymbol{\omega}^W
= \tfrac{1}{2}\big(\mathbf{c}^W\times \dot{\mathbf{c}}^W + \mathbf{s}^W\times \dot{\mathbf{s}}^W + \mathbf{n}^W\times \dot{\mathbf{n}}^W\big),
\label{eq:omega_world_general_fixed}
\end{equation}
which in the body frame is:$\boldsymbol{\omega}^B = R^\top\,\boldsymbol{\omega}^W.
$
Time derivatives are obtained by differentiating \eqref{eq:axes_world_wind}, noting that \(\mathbf{e}_a^W\) and \(\mathbf{n}_{\text{curve}}^W\) depend on \(\mathbf{v}_a^W(t)\) and \(\mathbf{F}_{\text{req}}^W(t)\).

\subsubsection*{Step 8: Aerodynamic moment coefficients.}
Define the required aerodynamic torque:
\begin{equation}
\boldsymbol{\tau}_{\text{req}}^B
:= I_B\,\dot{\boldsymbol{\omega}}^B+\boldsymbol{\omega}^B\times\!\big(I_B\boldsymbol{\omega}^B\big)
\;-\; \boldsymbol{\tau}_{\text{ext}}^B \;-\; \boldsymbol{\tau}_{\text{prop}}^B.
\label{eq:tau_req_wind}
\end{equation}
Assuming the linear aerodynamic moment model in~\eqref{eq:TaeroB} and using the Euler equation~\eqref{eq:rotation},
the coefficients are computed as:\footnote{Sign convention:
In this frame definition (\emph{robotic convention}), the \(y\)-axis \(\mathbf{e}_2^B\) points to the aircraft’s \emph{left}. Consequently, a positive pitching moment \(C_m>0\) creates a \emph{nose-down} rotation (right-hand rule around a left-pointing axis). In standard flight mechanics (FRD
), a positive pitching moment produces \emph{nose-up}, and a positive yawing moment is about the \emph{down}-axis. Care is therefore required when comparing \((C_\ell,C_m,C_n)\) computed here with FRD-based databases: \(C_m\) and \(C_n\) have opposite signs relative to those conventions, while \(C_\ell\) (roll about \(\mathbf{e}_1^B\)) is unaffected.}
\begin{equation}
C_\ell = \frac{\zeta\!\cdot\!\mathbf{e}_1^B}{ q\,S\,b,},\,
C_m   = 
\frac{\zeta\!\cdot\!\mathbf{e}_2^B}{
q\,S\,\bar c},\,
C_n   = \frac{(\zeta\!\cdot\!\mathbf{e}_3^B}{q\,S\,b},
\label{eq:moment_coeffs_wind}
\end{equation}
with $\zeta=\boldsymbol{\tau}_{\text{req}}^B - D_\omega\,\boldsymbol{\omega}^B.$

\subsubsection*{Step 9: Control Allocation.}
Finally, we solve for the physical control surface deflections \(\mathbf{u} = (\delta_a, \delta_e, \delta_r)^\top\).
While Step 5 decoupled \(\mathbf{u}\) from the translational forces, we now reintroduce the full model to satisfy the moment requirements.
Using a standard affine aerodynamic model:
\begin{equation}
\mathbf{C}_{\text{req}} = \mathbf{C}_{\text{passive}}(\alpha, \beta, \boldsymbol{\omega}^B) + \mathbf{B}_{\text{eff}}(\alpha)\,\mathbf{u},
\label{eq:control_allocation_model}
\end{equation}
where \(\mathbf{C}_{\text{req}}=(C_\ell, C_m, C_n)^\top\) are the coefficients from Step 8, \(\mathbf{C}_{\text{passive}}\) contains static stability and damping terms (independent of control), and \(\mathbf{B}_{\text{eff}}\) is the control effectiveness matrix.
The control inputs are found by linear inversion:
\begin{equation}
\mathbf{u} = \mathbf{B}_{\text{eff}}^{-1}(\alpha)\,\big( \mathbf{C}_{\text{req}} - \mathbf{C}_{\text{passive}} \big).
\label{eq:u_solver}
\end{equation}
\emph{Remark on coupling and feasibility:} If the control surfaces significantly affect lift and drag (e.g., large flaps or elevons), the approximation in Step 5 becomes inaccurate. In such cases, Steps 5–9 should be iterated: initialize with \(\mathbf{u}=\mathbf{0}\), solve for \(\alpha, T\), compute a new \(\mathbf{u}\), and update the aerodynamic forces in \eqref{eq:Faero_coord_wind} until convergence. If \(\mathbf{B}_{\text{eff}}\) is non-square or ill-conditioned, use a (regularized) pseudoinverse and enforce actuator limits.

\subsection{Degeneracies and edge cases}
\subsubsection*{Zero airspeed} If \(\|\mathbf{v}_a^W\|\approx 0\), the dynamic pressure vanishes (\(q\approx 0\)), making aerodynamic forces negligible and the air-relative frame ill-defined. In this case, thrust must provide the entire required force. Set \(\mathbf{c}^W = \mathbf{F}_{\text{req}}^W / \|\mathbf{F}_{\text{req}}^W\|\), choose \(\mathbf{n}^W\) to prevent roll singularity (e.g., maintain the previous attitude), and set \(T = \|\mathbf{F}_{\text{req}}^W\|\).

\subsubsection*{Zero perpendicular component} If \(f_\perp \approx 0\), the required force is purely axial (collinear with \(\mathbf{e}_a^W\)). The roll angle is dynamically undetermined. To maintain unique behavior (e.g., level flight), define \(\mathbf{n}_{\text{curve}}^W\) by projecting the world-up vector \(+\mathbf{e}_3^W\) onto the plane orthogonal to \(\mathbf{e}_a^W\). This aligns the lift axis with gravity, keeping the wings level.

\subsubsection*{Feasibility} Check bounds (stall angle \(\alpha_{\max}\), maximum thrust \(T_{\max}\), control saturation). If \eqref{eq:two_eqs_wind} cannot be met, the trajectory requires relaxation (e.g., reducing commanded acceleration or relaxing the \(\beta=0\) constraint).

\subsubsection*{Groundspeed vs airspeed} Strong winds may cause \(\mathbf{e}_a^W\) (heading) to deviate significantly from \(\mathbf{v}^W\) (course). Eq. \eqref{eq:decompose_wind} remains physically valid because aerodynamics fundamentally depends on the air-relative vector, not the inertial path.

\subsection{Approximate trim solution for simplified aerodynamics}
If the aerodynamic coefficients can be approximated as \(C_L(\alpha)\approx C_{L\alpha}\,\alpha\) and \(C_D(\alpha)\approx C_{D0}+k\,\alpha^2\), the implicit dependency in \eqref{eq:T_alpha_closed_wind} can be resolved by substituting \eqref{eq:T_components_wind} into the geometric relationship \(\tan\alpha = T_\perp/T_\parallel\):
\begin{equation}
\tan\alpha = \frac{f_\perp - q\,S\,C_{L\alpha}\,\alpha}{f_\parallel + q\,S\,(C_{D0}+k\,\alpha^2)}.
\label{eq:alpha_scalar_eq}
\end{equation}
This scalar equation for \(\alpha\) can be solved efficiently (e.g., via Newton’s method). The resulting \(\alpha^\star\) is then substituted back into \eqref{eq:T_alpha_closed_wind} and \eqref{eq:axes_world_wind} to obtain the corresponding thrust \(T(t)\) and orientation \(R(t)\). These remain \emph{pointwise quasi‑steady} solutions; they constitute a steady-flight \emph{trim} only when the trajectory and rotational dynamics are steady.

\section{IFD on a Spherical Parallel with External Force Aligned to the Sphere Center}
\label{sec:tethered}

We now particularize the inverse dynamics to a specific case study. We assume no wind, \(\mathbf{w}^W\equiv\mathbf{0}\), in order to isolate the influence of the CoM trajectory and force parameters on the attitude and control inputs.
All the assumptions, nomenclature, and given information stated in Sections~\ref{sec:model} and~\ref{sec:inverse} hold. 

We consider a desired CoM trajectory on the sphere of radius \(L\) centered at the origin, along a \emph{parallel of latitude} in the upper hemisphere: fix the colatitude\footnote{Colatitude \(\theta\) is the angle from \(+\mathbf{e}_3^W\); latitude \(\lambda=\frac{\pi}{2}-\theta\).}
\(\theta\in[0,\tfrac{\pi}{2}]\) (equivalently, height \(z_0=L\cos\theta\) and circle radius \(r=L\sin\theta\)),
and require a \emph{constant tangential speed} \(v_0>0\) along that circle.

Additionally, the external force \(\mathbf{f}_{\text{ext}}^W(t)\) has constant magnitude \(\|\mathbf{f}_{\text{ext}}^W(t)\|=F_{\text{ext}}\in\mathbb{R}\) and 
its direction is always radial, along the line connecting the CoM with the world-frame origin, as it would be if a cable were pulling the aircraft toward the center.
No external moment \(\boldsymbol{\tau}_{\text{ext}}^B(t)\) is present.

Let ${\omega_{\text{cir}}:=v_0/r>0,}\quad
\psi(t):=\psi_0+\omega_{\text{cir}}\,t.$
Then define 
\[
\mathbf{u}_{r\!xy}^W(t)=\begin{bsmallmatrix}\cos\psi\\ \sin\psi\\ 0\end{bsmallmatrix},\;
\mathbf{e}_t^W(t)=\begin{bsmallmatrix}-\sin\psi\\ \cos\psi\\ 0\end{bsmallmatrix},\;
\mathbf{e}_3^W=\begin{bsmallmatrix}0\\0\\1\end{bsmallmatrix}
\]
and note that 
\(\dot{\mathbf{u}}_{r\!xy}^W(t)=\omega_{\text{cir}}\,\mathbf{e}_t^W(t)\) and \(\dot{\mathbf{e}}_t^W(t)=-\omega_{\text{cir}}\,\mathbf{u}_{r\!xy}^W(t)\).
Then we have:
\begin{align}
\mathbf{p}^W(t)&=
r\, \mathbf{u}_{r\!xy}^W(t) + 
z_0\,\mathbf{e}_3^W,\\
\mathbf{v}^W(t)&=\dot{\mathbf{p}}^W(t)
= \omega_{\text{cir}}\, r \,\mathbf{e}_t^W(t)
= v_0\,\mathbf{e}_t^W(t),\\
\mathbf{a}^W(t)&=\dot{\mathbf{v}}^W(t)
= -\omega_{\text{cir}}\, v_0\,\mathbf{u}_{r\!xy}^W(t)
= -\frac{v_0^2}{r}\,\mathbf{u}_{r\!xy}^W(t).
\end{align}

Given the zero-wind assumption we have
\(\mathbf{e}_a^W=\mathbf{e}_t^W\), which implies
\(\mathbf{e}_a^W\perp\mathbf{u}_{r\!xy}^W\), \(\dot{\mathbf{e}}_a^W=-\omega_{\text{cir}}\mathbf{u}_{r\!xy}^W\), and \(\dot{\mathbf{u}}_{r\!xy}^W=\omega_{\text{cir}}\mathbf{e}_a^W\).

\subsubsection*{External force (radial to origin).}

At the point \(\mathbf{p}^W(t)\) on the sphere, the outward radial unit is 
\[
\mathbf{e}_r^W(t)
=({r}/{L})\,\mathbf{u}_{r\!xy}^W+({z_0}/{L})\,\mathbf{e}_3^W,
\]
thus the external force is written as\footnote{The minus sign has been chosen such that \(F_{\text{ext}}>0\) implies an external force that pulls the aircraft toward the center (inside the sphere), e.g., the one produced by a cable attaching the aircraft CoM to the origin of \(W\). In order to balance such a force and stay on the trajectory the aircraft must produce a force that pulls the cable outward, outside the sphere.}
\[
\mathbf{f}_{\text{ext}}^W(t)=-F_{\text{ext}}\,\mathbf{e}_r^W(t)
=-\frac{F_{\text{ext}}\,r}{L}\,\mathbf{u}_{r\!xy}^W \;-\; \frac{F_{\text{ext}}\,z_0}{L}\,\mathbf{e}_3^W.
\]

\subsubsection*{Required net (non-gravitational) force }
Plugging the acceleration, gravity, and external-force expressions into~\eqref{eq:F_req_wind} we obtain the required world-frame force to be generated by propulsion+aerodynamics for the particular case at hand:
\begin{equation}
\mathbf{F}_{\text{req}}^W
= \underbrace{\Big(-\,m\,\tfrac{v_0^2}{r} + \tfrac{F_{\text{ext}}\,r}{L}\Big)}_{A_h}\,\mathbf{u}_{r\!xy}^W
\;+\;
\underbrace{\Big(mg + \tfrac{F_{\text{ext}}\,z_0}{L}\Big)}_{A_z}\,\mathbf{e}_3^W,
\label{eq:F_req_components_fixed}
\end{equation}
where \(A_h\) and \(A_z\) are constant. 
Since \(\mathbf{F}_{\text{req}}^W\perp \mathbf{e}_a^W\), the longitudinal component is zero:
\(f_\parallel=\mathbf{F}_{\text{req}}^W\cdot\mathbf{e}_a^W=0\).
The direction and norm of the perpendicular component are
\begin{equation}
\mathbf{n}_{\text{curve}}^W
= {\mathbf{F}_{\text{req}}^W}/{\|\mathbf{F}_{\text{req}}^W\|}
= ({A_h\,\mathbf{u}_{r\!xy}^W + A_z\,\mathbf{e}_3^W})/{\sqrt{A_h^2 + A_z^2}},
\label{eq:ncurve_fixed_dir}
\end{equation}
\begin{equation}
f_\perp=\big\|\mathbf{F}_{\text{req}}^W\big\|=\sqrt{A_h^2 + A_z^2}.
\label{eq:ncurve_fixed_norm}
\end{equation}

The choice that enforces \emph{coordinated flight} (\(\beta=0\)) is
\begin{equation}
\mathbf{s}^W = \mathbf{n}_{\text{curve}}^W \times \mathbf{e}_a^W 
= \tfrac{A_h}{\sqrt{A_h^2 + A_z^2}}\,\mathbf{e}_3^W
\;-\; \tfrac{A_z}{\sqrt{A_h^2 + A_z^2}}\,\mathbf{u}_{r\!xy}^W,
\label{eq:s_world_fixed}
\end{equation}
so \(\{\mathbf{e}_a^W,\mathbf{s}^W,\mathbf{n}_{\text{curve}}^W\}\) is a right-handed orthonormal triad which acts as an intermediate frame. The final attitude is derived by such a frame after a rotation about \(\mathbf{s}^W\) of the angle of attack \(\alpha\), which is among the next quantities to be found.

\subsubsection*{Thrust alignment and implicit equations for \(T\) and \(\alpha\)}

The dynamic pressure \(q=\tfrac{1}{2}\rho v_0^2\) is \emph{constant}.
With \(\beta=0\), the aerodynamic force is given by~\eqref{eq:Faero_coord_wind}.

Substituting \(f_\parallel=0\) and \(f_\perp = \sqrt{A_h^2 + A_z^2}\) into~\eqref{eq:two_eqs_wind}, and then using~\eqref{eq:T_alpha_closed_wind}, we obtain:
\begin{equation}
\begin{aligned}
T &= \sqrt{(qS\,C_D)^2 + \big(\sqrt{A_h^2 + A_z^2}-qS\,C_L\big)^2},\\
\alpha &= \atanTwo\!\big(\sqrt{A_h^2 + A_z^2}-qS\,C_L,\; qS\,C_D\big).
\end{aligned}
\label{eq:T_alpha_closed_fixed}
\end{equation}
Thus,~\eqref{eq:T_alpha_closed_fixed} constitutes the implicit system to be solved for \((T,\alpha)\).
Note that \(\sqrt{A_h^2+A_z^2}\) is constant along the circle.

\subsubsection*{Aircraft attitude and angular velocity.}

Once \(\alpha\) is found, compute \(\mathbf{c}^W\) and \(\mathbf{n}^W\) by rotating about \(\mathbf{s}^W\) using~\eqref{eq:axes_world_wind}, which, assembled in~\eqref{eq:R_closed_wind}, give the rotation matrix \(R(t)\) that represents the aircraft attitude.
With constant \(v_0\) and fixed maps evaluated at \(\alpha\), \(\alpha\) is constant and \(R(t)\) executes a uniform rotation about \(\mathbf{e}_3^W\) at rate \(\omega_{\text{cir}}\).

The angular velocity associated with \(R(t)\) is computed using~\eqref{eq:omega_world_general_fixed}. 
Given that \(\dot{\mathbf{e}}_a^W=-\omega_{\text{cir}}\mathbf{u}_{r\!xy}^W\), \(\dot{\mathbf{u}}_{r\!xy}^W=\omega_{\text{cir}}\mathbf{e}_a^W\), \(\dot{\mathbf{e}}_3^W=\mathbf{0}\), and \(\alpha\) is constant, one obtains a \emph{constant} \(\boldsymbol{\omega}^W\) parallel to \(\mathbf{e}_3^W\): $
\boldsymbol{\omega}^W(t) = \omega_{\text{cir}}\,\mathbf{e}_3^W,\qquad
\boldsymbol{\omega}^B(t) = R(t)^\top\,\boldsymbol{\omega}^W(t).
$

\subsubsection*{Aerodynamic moment coefficients.}

The aerodynamic moment coefficients do not have a special structure in this case and are computed using~\eqref{eq:moment_coeffs_wind}.

\subsection{Trim Solution \texorpdfstring{$(T,\alpha)$}{(T,alpha)} under a Simple Lift/Drag Law}
\label{sec:trim_solution}

Assume the small-angle law \(C_L(\alpha)\approx a\,\alpha\), \(C_D(\alpha)\approx C_{D0}+k_\alpha\,\alpha^2\), \(\tan\alpha\approx\alpha\), \(\sin\alpha\approx\alpha\), and \(\cos\alpha\approx 1\). Let \(Q:=qS\) and remember that in this case \(f_\parallel =0\) and \(f_\perp=\sqrt{A_h^2+A_z^2}\) (constant).
From the axial/normal balance~\eqref{eq:two_eqs_wind},
with the aerodynamic model shorthand \(F_L=Q\,a\alpha\), \(D=Q\,(C_{D0}+k_\alpha\,\alpha^2)\),
we have \(T\cos\alpha=D\) (i.e., \(T\approx D\)), and \(T\sin\alpha+F_L=f_\perp\) (i.e., \(T\,\alpha + F_L \approx f_\perp\)). Substituting the first into the second we obtain \(D\,\alpha + F_L \approx f_\perp\). Substituting back the aerodynamic models leads to a cubic equation for the trim angle of attack:
\begin{equation}
k_\alpha\,\alpha^3 + (C_{D0} + a)\,\alpha = {f_\perp}/{Q}.
\label{eq:cubic_trim}
\end{equation}
This is solved using Cardano’s formula for the unique real root \(\alpha\) (for typical aerodynamic parameters \(k_\alpha>0\) and \(C_{D0}+a>0\), the left-hand side is strictly increasing, hence the solution is unique). Let \(p=\frac{C_{D0}+a}{k_\alpha}\), \(d=\frac{f_\perp}{Qk}\), \(\Delta=\big(\tfrac{d}{2}\big)^2+\big(\tfrac{p}{3}\big)^3\), and \(u_\pm=\tfrac{d}{2}\pm\sqrt{\Delta}\).
Then 
$
\alpha = \sqrt[3]{u_+} + \sqrt[3]{u_-}.
\label{eq:alpha_trim_closed}
$
The required thrust is then obtained directly from the drag:
\begin{equation}
T = {Q\,(C_{D0} + k_\alpha\,\alpha^2)}/{\cos\alpha}.
\label{eq:T_trim_closed}
\end{equation}
We recall that since \(v_0\), \(r\), \(L\), and \(F_{\text{ext}}\) are constant, then \(f_\perp\) is constant, and thus the trim solution \((T,\alpha)\) is constant in time.


\subsection{Bank Dependence on External Force and Geometry}
\label{sec:cable_bank_dependence}

In this section we study the dependence of the constant bank angle on the external force intensity and the geometric parameters of the desired trajectory.
It is convenient to define the \emph{inward} horizontal demand
\[
A_h^{\text{in}} := -A_h = m\,\frac{v_0^2}{r} - F_{\text{ext}}\,\frac{r}{L},
\quad
A_z := mg + F_{\text{ext}}\,\frac{z_0}{L}.
\]
The signed geometric bank angle (positive inward) obeys
\begin{equation}
\tan\mu = {A_h^{\text{in}}}/{A_z}
= 
({\tfrac{v_0^2}{rg} - F_{\text{ext}}\,\tfrac{r}{Lmg}})/
({1 + F_{\text{ext}}\,\tfrac{z_0}{Lmg}})
.
\label{eq:bank_signed_fixed}
\end{equation}
Introducing the dimensionless
\(\kappa := \frac{v_0^2}{gL}\) and \(\eta := \frac{F_{\text{ext}}}{mg}\) (normalized external force),
and substituting the expressions of \(r\) and \(z_0\) as functions of \(L\) and \(\theta\), one obtains
\begin{equation}
\tan\mu
= ({\displaystyle {\kappa}/{\sin\theta}\;-\;\eta\,\sin\theta})/
       ({\displaystyle 1\;+\;\eta\,\cos\theta})\ .
\label{eq:bank_theta_fixed}
\end{equation}
This makes the force/speed/latitude dependencies explicit and matches the classical relation \(\tan\mu=\frac{v_0^2}{g\,r}\) when \(\eta=0\).

\subsubsection*{Geometric Nature of the Bank Angle}
The expression for the bank angle \(\mu\) derived in~\eqref{eq:bank_theta_fixed} depends exclusively on the trajectory parameters (\(v_0, r\)), the inertial properties (\(m, g\)), and the external force magnitude (\(F_{\text{ext}}\)). It is independent of the aircraft’s aerodynamic properties (such as \(C_L, C_D\), wing area \(S\), or air density \(\rho\)) thanks to the kinematic constraint of \emph{coordinated flight} (\(\beta=0\)): the aircraft geometry decouples, forcing the airframe to align with the resultant force vector determined solely by Newton’s laws applied to the CoM.

\subsection{Locus of Parameters for the Efficient Zero Bank Attitude}
\label{sec:zero_bank_locus}

The zero-bank condition (\(\mu=0\)) represents a “sweet spot” for efficiency.
This claim can be justified analytically using the force balance and drag polar introduced in Section~\ref{sec:trim_solution}.
The total required perpendicular force is \(f_\perp = F_L + T\sin\alpha\). For efficient cruising flight where \(\alpha \ll 1\), the thrust contribution is negligible, implying \(F_L \approx f_\perp = \sqrt{A_h^2 + A_z^2}\).
Using the parabolic drag model \(C_D = C_{D0} + k\alpha^2\) and the linear lift model \(C_L = a\alpha\), the total drag force \(D\) can be expressed as a function of lift:
\begin{equation}
D(F_L) = qS\,C_{D0} + F_L^2 ({k}/{qS\,a^2})\,.
\label{eq:drag_lift_relation}
\end{equation}
Since the dynamic pressure \(q\) is constant, the parasitic drag term is fixed. The induced drag scales with \(F_L^2 \approx f_\perp^2 = A_h^2 + A_z^2\).
At the zero-bank equilibrium, the horizontal demand vanishes (\(A_h=0\)), minimizing \(f_\perp\) to its lower bound \(|A_z|\). Any deviation from zero bank introduces a non-zero \(A_h\), increasing \(f_\perp\) and so increasing the induced drag quadratically.

Therefore, we seek the condition under which the aircraft maintains a level-wings attitude (\(\mu=0\)) despite the curvilinear motion and external tether force. This corresponds to the case where the net side force in the body frame is zero purely via the balance of centrifugal and external forces, without requiring aerodynamic banking.
Setting the numerator of~\eqref{eq:bank_theta_fixed} to zero yields the \emph{zero-bank condition}:
\begin{equation}
{\kappa}/{\sin\theta} - \eta\,\sin\theta = 0
\quad\Longrightarrow\quad
\eta = {\kappa}/{\sin^2\theta}.
\label{eq:zero_bank_condition}
\end{equation}
In physical variables, substituting \(\eta=\frac{F_{\text{ext}}}{mg}\) and \(\kappa=\frac{v_0^2}{gL}\), this becomes $
\frac{F_{\text{ext}}\,r}{L} = \frac{m v_0^2}{r}.$
This recovers the intuitive result that zero bank is achieved when the horizontal projection of the external force exactly balances the centrifugal force \(m v_0^2/r\).

\subsection{Sensitivity Analysis at the zero-bank locus}
\label{sec:sensitivity}

We examine how the bank angle \(\mu\) changes for small deviations in the parameters \((\eta, \kappa, \theta)\) around the zero-bank equilibrium defined by~\eqref{eq:zero_bank_condition}.
Let the right-hand side of~\eqref{eq:bank_theta_fixed} be \(f(\eta,\kappa,\theta) = N/\mathcal{D}\), 
where $
N := {\kappa}/{\sin\theta} - \eta\,\sin\theta$, 
$\mathcal{D} := 1 + \eta\,\cos\theta.$
Since \(\mu=0\) at the locus, \(\frac{\partial \mu}{\partial (\cdot)} \approx \frac{\partial \tan\mu}{\partial (\cdot)}\).
Furthermore, since the numerator \(N=0\) at the locus, the quotient rule simplifies to
\(\frac{\partial \mu}{\partial x} \approx \frac{1}{\mathcal{D}}\,\frac{\partial N}{\partial x}\).

Evaluating the partial derivatives at the locus \(\eta = \kappa/\sin^2\theta\):

\paragraph*{Sensitivity to External Force (\(\eta\))}
\[
{\partial \mu}/{\partial \eta}\big|_{\mu=0}
= {-\sin\theta}/({1 + \eta\cos\theta)}.
\]
Since the denominator is positive (assuming \(F_{\text{ext}}\) does not push the aircraft inverted), the derivative is \emph{negative}. Increasing the inward tether force \(F_{\text{ext}}\) pulls the aircraft “outward” (negative \(\mu\)), reducing the need for aerodynamic banking into the turn.

\paragraph*{ Sensitivity to Speed (\(\kappa\))}
\[
{\partial \mu}/{\partial \kappa}\big|_{\mu=0}
=1/({\sin\theta\,\big(1 + \eta\cos\theta\big)}).
\]
The derivative is \emph{positive}. Increasing speed increases centrifugal force, requiring a positive (inward) bank angle to compensate.

\begin{figure*}[t]
    \centering
    \begin{subfigure}[b]{0.32\textwidth}
        \centering
        \includegraphics[width=\linewidth]{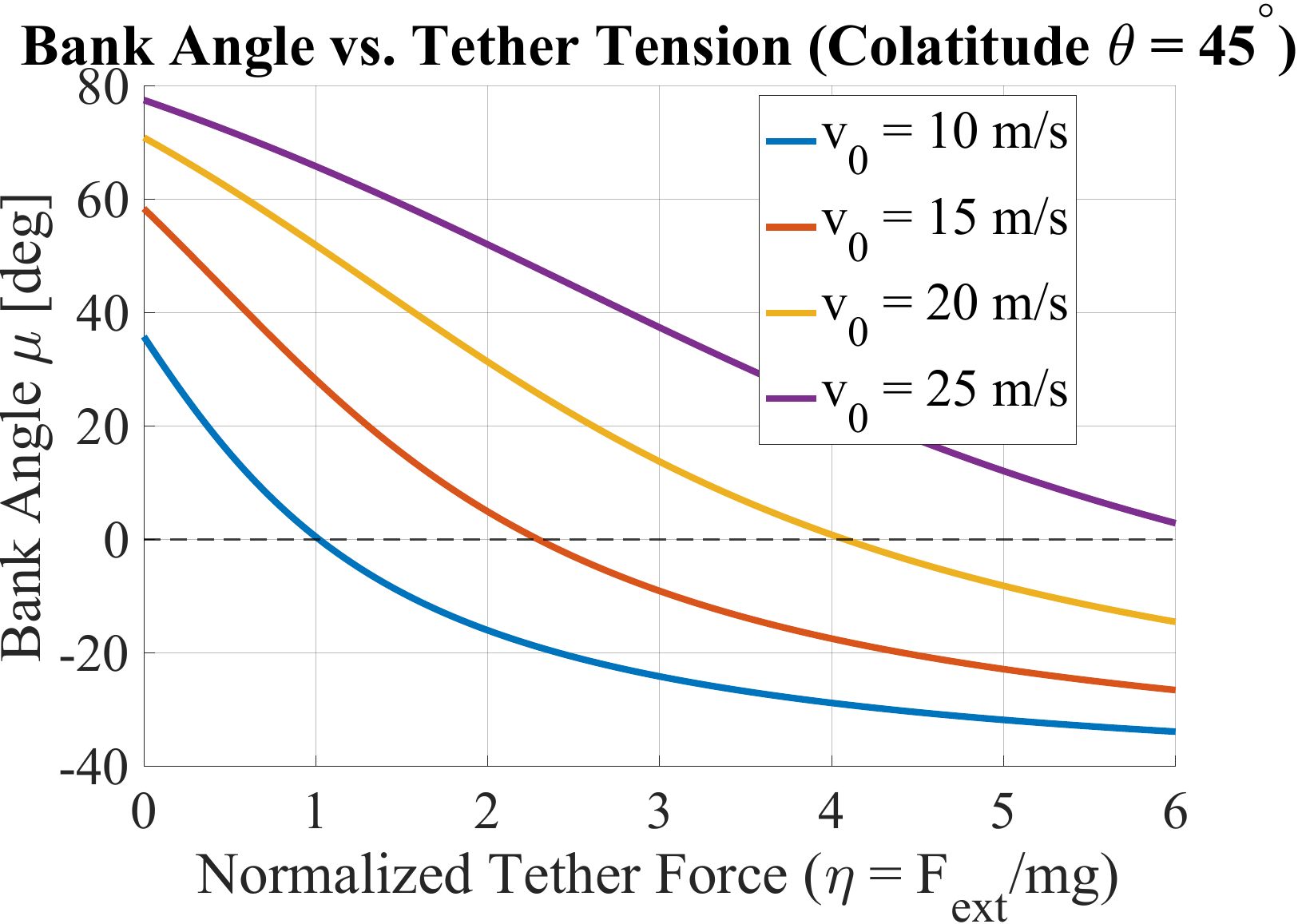}
        \caption{Sensitivity to Tether Force}
        \label{fig:bank_force}
    \end{subfigure}
    \hfill
    \begin{subfigure}[b]{0.32\textwidth}
        \centering
        \includegraphics[width=\linewidth]{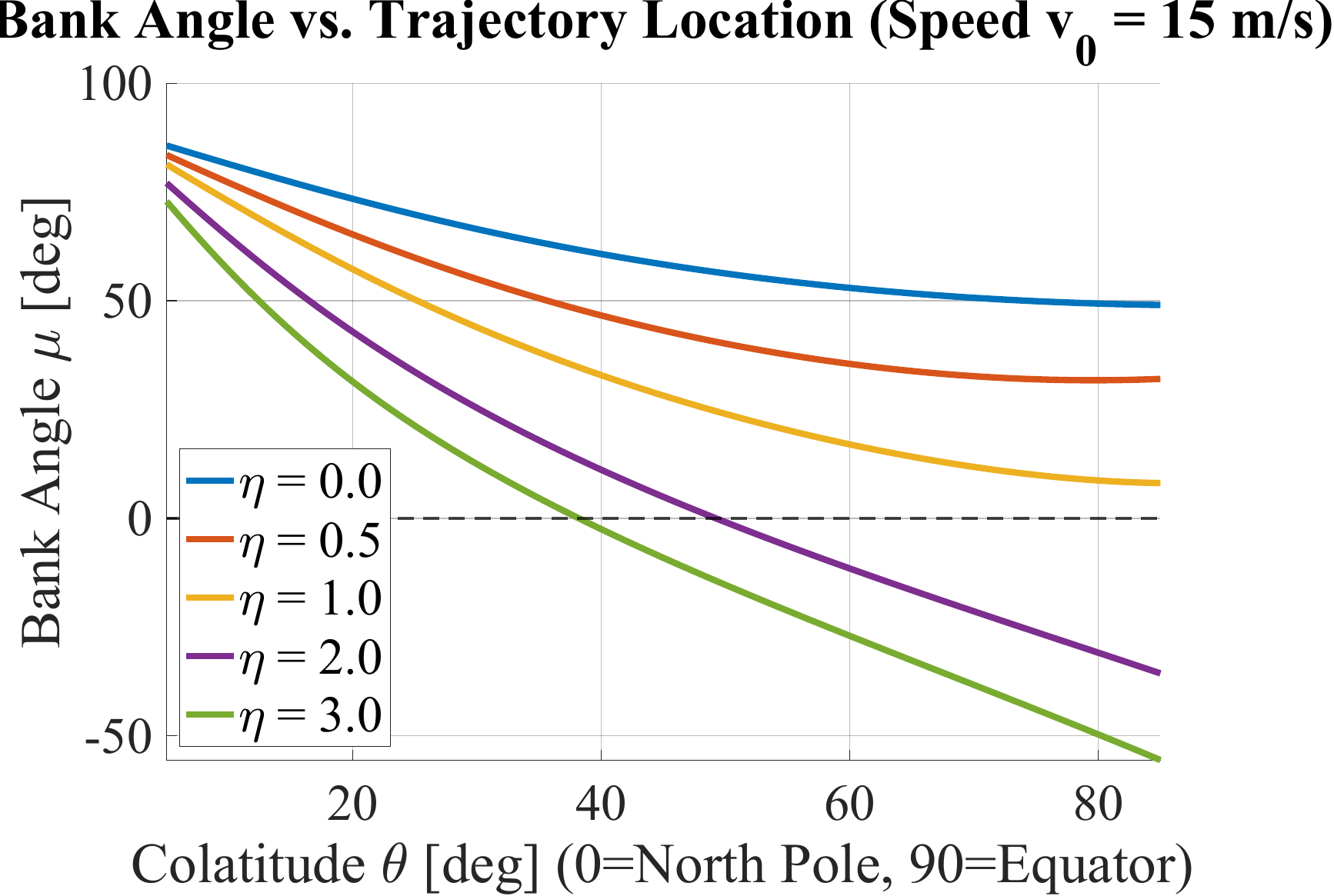}
        \caption{Sensitivity to Trajectory Latitude}
        \label{fig:bank_latitude}
    \end{subfigure}
    \hfill
    \begin{subfigure}[b]{0.32\textwidth}
        \centering
        \includegraphics[width=\linewidth]{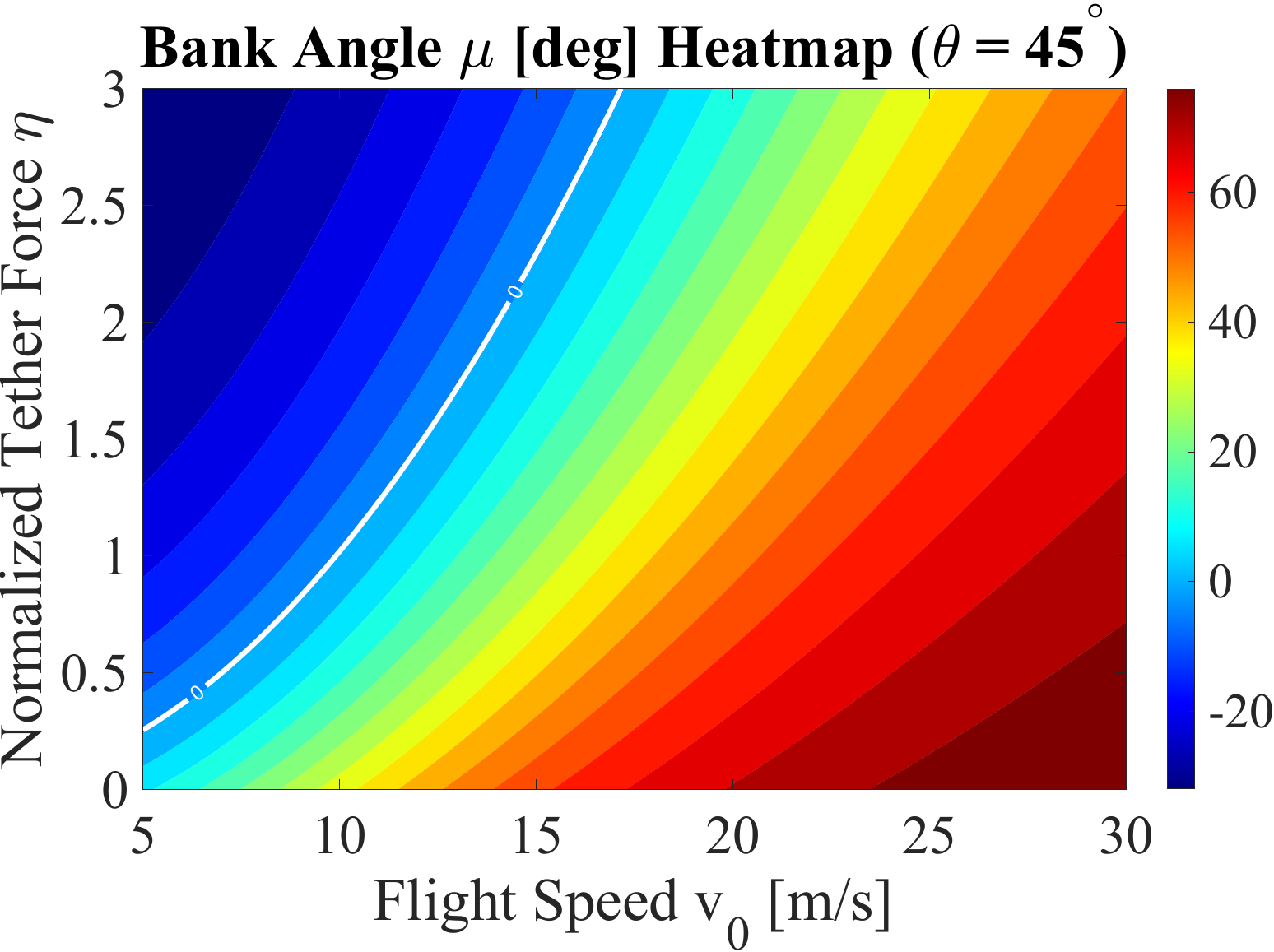}
        \caption{zero-bank locus (White Line)}
        \label{fig:bank_heatmap}
    \end{subfigure}
    \caption{Behavior of the geometric bank angle \(\mu\) required for coordinated tethered flight. \textbf{(a)} Increasing tether tension pulls the aircraft outward (negative bank), competing with centrifugal force. \textbf{(b)} Lowering the trajectory on the sphere (increasing \(\theta\)) increases the turn radius, reducing centrifugal force and favoring outward banking. \textbf{(c)} The heatmap highlights the zero-bank locus (white contour), separating the regime dominated by centrifugal force (positive bank) from the regime dominated by tether tension (negative bank).}
    \label{fig:bank_analysis}
\end{figure*}

\paragraph*{ Sensitivity to Trajectory Geometry (\(\theta\))}
The derivative of the numerator with respect to \(\theta\) is:
$\frac{\partial N}{\partial \theta} = -\frac{\kappa\cos\theta}{\sin^2\theta} - \eta\cos\theta.$
Substituting \(\eta = \kappa/\sin^2\theta\), we get \(-\frac{2\kappa\cos\theta}{\sin^2\theta}\). Thus:
\[
\partial \mu/{\partial \theta}\big|_{\mu=0}
= {-2\kappa\cos\theta}/({\sin^2\theta\,\big(1 + \eta\cos\theta\big)}).
\]
For \(\theta \in (0, \pi/2)\), this is \emph{negative}. Moving the trajectory “down” the sphere (increasing \(\theta\)) increases the radius \(r\), reducing centrifugal force and increasing the horizontal leverage of the tether, both of which push the bank angle negative.

\paragraph*{Sensitivity to Tether Length (\(L\))}
The parameter \(L\) appears implicitly in \(\kappa = v_0^2/(gL)\).
Assuming fixed physical speed \(v_0\) and force \(F_{\text{ext}}\), we have \(\frac{\partial \kappa}{\partial L} = -\frac{\kappa}{L}\).
Applying the chain rule \(\frac{\partial \mu}{\partial L} = \frac{\partial \mu}{\partial \kappa}\frac{\partial \kappa}{\partial L}\):
\[
{\partial \mu}/{\partial L}\big|_{\mu=0}
= {-\kappa}/({L\,\sin\theta\,\big(1 + \eta\cos\theta\big)}).
\]
The derivative is \emph{negative}. Increasing the tether length \(L\) reduces the dimensionless speed parameter \(\kappa\) (and thus the centrifugal force), which shifts the balance toward the tether tension, inducing a negative (outward) bank tendency.

The results of the sensitivity analysis are summarized in Table~\ref{tab:sensitivity_summary}. This qualitative overview highlights how each physical parameter influences the aircraft’s equilibrium bank angle, confirming that tether tension and speed act as opposing drivers for the bank requirement.

\subsection{Behavior of the Bank Angle Curves}
\label{sec:bank_curves}

Analyzing the global behavior of~\eqref{eq:bank_theta_fixed} reveals three distinct regimes governed by the normalized force \(\eta\):

\begin{enumerate}[leftmargin=*]
\item  \textbf{Low Force / High Speed (\(\eta < \kappa/\sin^2\theta\)):}
  Here \(\mu > 0\). The centrifugal force dominates. The aircraft must bank \emph{inward} (into the turn) like a conventional aircraft, though less than it would without the tether.
    
    \item \textbf{Zero Bank (\(\eta = \kappa/\sin^2\theta\)):}
    The aircraft wings are level. The tether provides the required centripetal acceleration.
    
    \item \textbf{High Tension / Low Speed (\(\eta > \kappa/\sin^2\theta\)):}
    Here \(\mu < 0\). The tether pull is stronger than the centrifugal force. The aircraft must bank \emph{outward} (away from the turn center) to generate aerodynamic lift that opposes the tether force.
\end{enumerate}


\begin{table}[b]
\centering
\caption{Sensitivity of Bank Angle at the Zero-Bank Equilibrium}
\label{tab:sensitivity_summary}
\renewcommand{\arraystretch}{1.2}
\begin{tabularx}{\linewidth}{>{\hsize=.7\hsize}X c >{\hsize=1.3\hsize}X}
\hline
\textbf{Symbol \& Parameter} & \textbf{Sign of} \(\frac{\partial \mu}{\partial (\cdot)}\) & \textbf{Physical Interpretation} \\
\hline
\(\eta\): Normalized External Force & (-) & Higher tether tension requires the aircraft to bank outward. \\
\(\kappa\): Dimensionless Speed Squared & (+) & Higher speed increases centrifugal force, requiring inward bank. \\
\(\theta\): Colatitude & (-) & Increasing \(\theta\) (moving down) increases radius, reducing centrifugal force. \\
\(L\): Tether Length & (-) & Longer tether reduces angular rate (for fixed \(v_0\)), thus centrifugal force. \\
\hline
\end{tabularx}
\end{table}

\subsubsection*{Asymptotic Limit.}
As the tether force becomes very large:
$
\lim_{\eta\to\infty} \tan\mu = -\tan\theta
$, the bank angle $ \mu$ approaches $-\theta$.
Geometrically, this implies the aircraft body \(z\)-axis aligns with the tether radius vector \(\mathbf{e}_r^W\). In this limit, the wing plane is perpendicular to the tether, and so lift vector points opposite to the tether tension, which is the expected equilibrium for a kite or tethered drone dominated by cable tension.

A comprehensive visualization of these regimes is provided in Figure~\ref{fig:bank_analysis}. The plots illustrate the transition from positive to negative bank angles as the tether force increases (Fig.~\ref{fig:bank_analysis}a) or as the trajectory moves to lower latitudes (Fig.~\ref{fig:bank_analysis}b), with the specific zero-bank boundary explicitly mapped in Figure~\ref{fig:bank_analysis}c.

\emph{Remark [Independence from platform specifics.]}
All the results and plots in this subsection are \emph{platform‑agnostic}. The bank‑angle law is written entirely in terms of the \emph{dimensionless} parameters
\(\eta\) and \(\kappa\), together with the geometric colatitude \(\theta\).
Because forces are normalized by \(m g\) and speeds by \(\sqrt{gL}\), the curves do \emph{not} depend on aerodynamic properties
(\(S\), \(C_{L\alpha}\), \(C_{D0}\), \(k\)) nor on inertial/mechanical details (e.g., mass distribution \(I_B\), actuator limits).
In this sense, the bank angle \(\mu\) shown in Fig.~\ref{fig:bank_analysis} is a purely \emph{kinematic} consequence of:
(i) the spherical geometry, (ii) the radial external load, and (iii) the coordinated‑flight constraint \((\beta=0)\).
Platform‑specific parameters only re‑enter the analysis when the required net force is mapped to trim variables \((T,\alpha)\) or to aerodynamic moments and control inputs.

\section{Numerical Results}

\begin{figure*}[t]
    \includegraphics[trim={0cm 0cm 15cm 5cm},clip, width=0.3\textwidth]{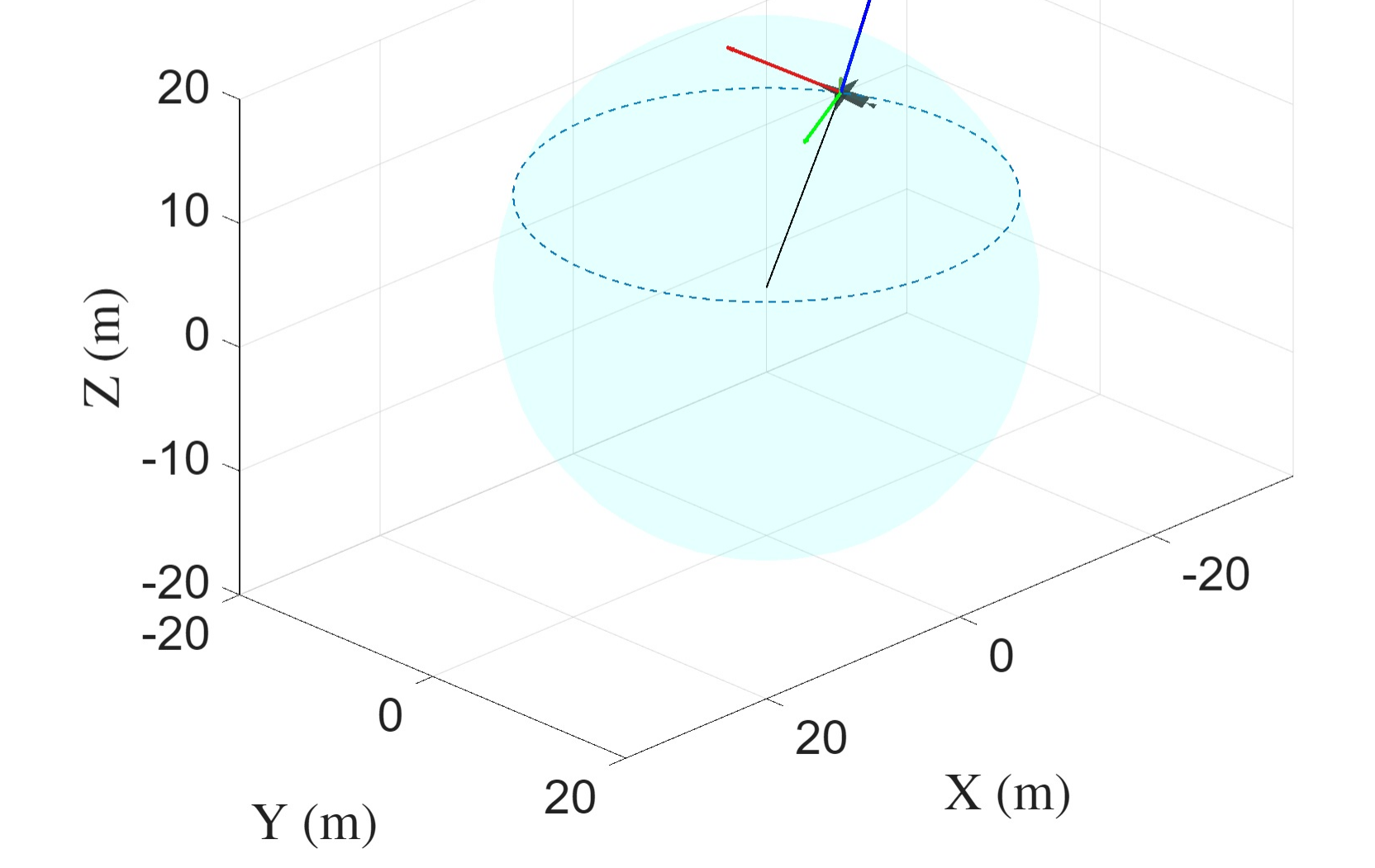}\includegraphics[width=0.35\textwidth]{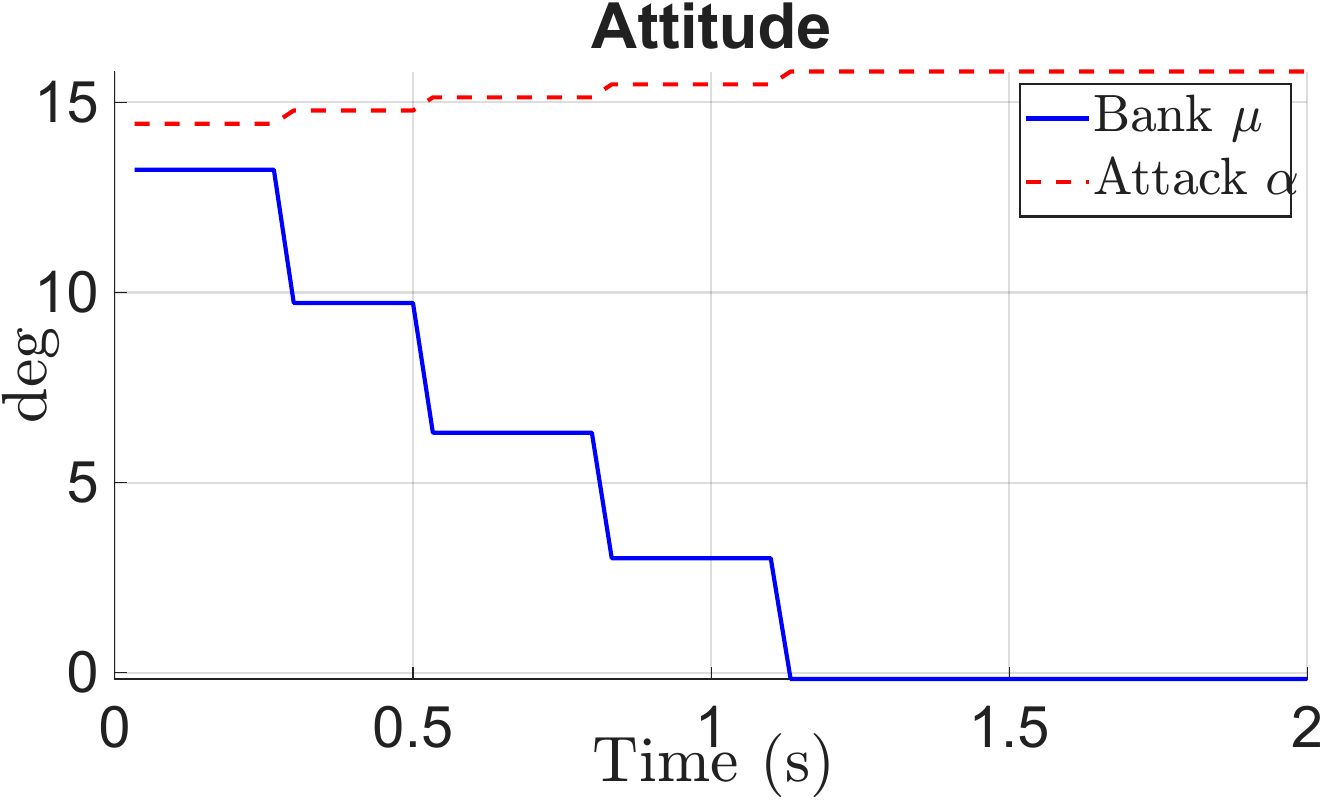}\includegraphics[width=0.35\textwidth]{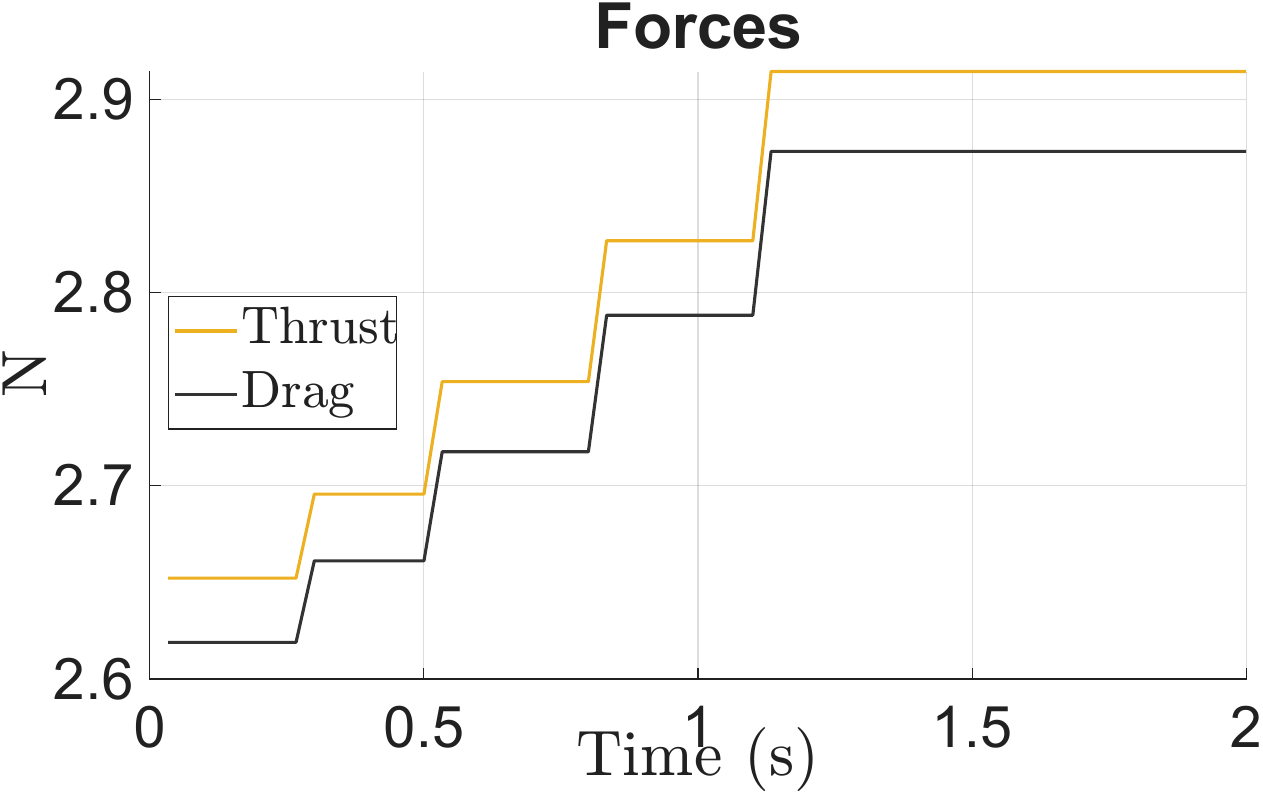}
    \caption{Attitude,  thrust, and drag forces for the aircraft schematically depicted on the left. The constant tangential speed is $11.7\rm{m/s}$ on a circle of radius $18.544\rm{m}$. The cable tension increases from $10\rm{N}$ to $16\rm{N}$ with steps of $1.5\rm{N}$, and the quantities in the displayed plots vary accordingly. For a cable tension of $16\rm{N}$, $\mu$ approaches zero, namely, the cable force compensates the centrifugal effects.}\label{fig:sim}
\end{figure*}

We present numerical results where the attitude and thrust of a tethered aircraft are computed using the proposed closed‑form solution. We use the ISA sea‑level air density \(\rho=1.225\,\mathrm{kg/m^3}\) and parameters for a \(2\,\mathrm{kg}\) aircraft chosen within realistic ranges \ifarxiv (Table~\ref{tab:sim_params} of the supplementary material).\else (Table IV of the supplementary material at \url{http://arxiv.org/abs/2602.17166}).\fi Specifically, the wing area is \(S=0.25\,\mathrm{m^2}\), the aspect ratio is \(\mathrm{AR}=5.6\), 
and the Oswald efficiency factor is \(e=0.8\). The aerodynamic coefficients are \(C_{L\alpha}=4.3\,\mathrm{rad^{-1}}\) and \(C_{D0}=0.035\). The cable length is \(L=20\,\mathrm{m}\); the colatitude is \(\theta=60^\circ\), yielding a circular‑trajectory radius of \(18.544\,\mathrm{m}\); and the constant tangential speed is \(v_0=11.7\,\mathrm{m/s}\). 
Figure~\ref{fig:sim} reports attitude and thrust for different cable tensions. When the cable tension balances the centrifugal force, the bank angle approaches zero. Under these conditions, the angle of attack is \(\simeq 16^\circ\). These results were obtained with standard parameters for small‑scale aircraft not designed to operate with a cable in this setting. To achieve a lower angle of attack, an aircraft should be designed to operate at a lower speed while producing more lift at that speed (e.g., a larger wingspan); this is outside the scope of the present paper. 

A wider range of tangential speeds, cable lengths, colatitude values, and cable tensions is illustrated in the accompanying video\footnote{\url{https://codeberg.org/CGabelli/Inverse-Flight-Dynamics-Tethered-Fixed-Wing-Aircraft}}. 
 For reproducibility, the software to generate the simulations, allowing interactive selection of the parameters, is available at the same link.

\section{Conclusion}
\label{sec:conclusion}

We presented a coordinate-free, 
formulation of fixed-wing flight dynamics on \(\SO\), enforcing the coordinated flight constraint to derive closed-form, pointwise quasi‑steady inversion solutions for thrust, angle of attack, and attitude. Applied to a tethered aircraft tracing spherical parallels, the method yielded analytic expressions for the bank angle and identified the \emph{zero-bank locus} where tether tension perfectly balances centrifugal forces. 
A sensitivity analysis revealed the competing influence of flight speed and tether tension on the bank requirement, while the structural analysis highlighted a fundamental decoupling between aerodynamic coordination (zero sideslip) and the perceived specific force.

Future work will extend the approach to wind effects and gusts, flexible cable dynamics, and experimental validation on fixed‑wing platforms, and will further clarify the mapping between the robotics FLU convention and standard aeronautics FRD moment signs when comparing databases.

\printbibliography[title={References}]
\end{refsection}

\ifarxiv
\clearpage 
    
    \twocolumn[
        \begin{center}
            \LARGE \textbf{Supplementary Material}{\\[1em]} 
            \large \mytitle{\\[1em]}
            \normalsize Antonio Franchi and Chiara Gabellieri{\\[2em]}
        \end{center}
    ]
    
    \appendices
    \normalfont

\begin{refsection}

\section{Auxiliary Tables}

For the reader's convenience, Tables~\ref{tab:symbols} and~\ref{tab:model_summary} provide a summary of the nomenclature and key model equations used in the main body of this paper.

\begingroup
\setlength{\tabcolsep}{3pt}
\renewcommand{\arraystretch}{1.20}
\begin{table*}[tb!]
\small
\caption{Symbols, units, and frames used throughout the geometric \( \SO \) flight-dynamics model. The frame in which vectors are expressed is specified in the corresponding column.}
\label{tab:symbols}

\begin{minipage}[t]{0.49\textwidth}
\centering
\begin{tabular}{llcl}
\hline
\textbf{Symbol} & \textbf{Description} & \textbf{Units} & \textbf{Frame} \\
\hline
\(\mathbf{p}^W \in \mathbb{R}^3\) & CoM position & m & \(W\) \\
\(\mathbf{v}^W \in \mathbb{R}^3\) & CoM velocity (\(\dot{\mathbf{p}}^W\)) & m/s & \(W\) \\
\(R \in \mathrm{SO}(3)\) & Rotation matrix (\(B \to W\)) & – & – \\
\(\boldsymbol{\omega}^B \in \mathbb{R}^3\) & Ang. vel. of \(B\) w.r.t \(W\) & rad/s & \(B\) \\
\(m > 0\) & Mass & kg & – \\
\(I_B \in \mathbb{R}^{3\times3}\) & Inertia (pos-def, const.) & kg·m\(^2\) & \(B\) \\
\(\mathbf{g}^W \in \mathbb{R}^3\) & Gravity vector & m/s\(^2\) & \(W\) \\
\(\mathbf{w}^W \in \mathbb{R}^3\) & Wind velocity & m/s & \(W\) \\
\(\mathbf{v}_a^W \in \mathbb{R}^3\) & Air-relative velocity & m/s & \(W\) \\
\(\mathbf{v}_a^B \in \mathbb{R}^3\) & Air-relative velocity & m/s & \(B\) \\
\(\mathbf{e}_a^B \in \mathbb{S}^2\) & Flow direction (unit) & – & \(B\) \\
\(\rho\) & Air density & kg/m\(^3\) & – \\
\(q\) & Dynamic pressure & Pa & – \\
\(S\) & Reference area & m\(^2\) & – \\
\(\bar c\) & Mean aero. chord & m & – \\
\(b\) & Wingspan & m & – \\
\(\mathbf{c}^B,\mathbf{s}^B,\mathbf{n}^B\) & Chord, span, normal units & – & \(B\) \\
\(\mathbf{e}_D^B,\mathbf{e}_L^B,\mathbf{e}_Y^B\) & Drag, lift, side units & – & \(B\) \\
\(\alpha\) & Angle of attack & rad & – \\
\hline
\end{tabular}
\end{minipage}
\hfill
\begin{minipage}[t]{0.49\textwidth}
\centering
\begin{tabular}{llcl}
\hline
\textbf{Symbol} & \textbf{Description} & \textbf{Units} & \textbf{Frame} \\
\hline
\(\beta\) & Sideslip angle & rad & – \\
\(\mathbf{u}\) & Control inputs & – & – \\
\(\mathbf{u}_t^B \in \mathbb{S}^2\) & Thrust direction (unit) & – & \(B\) \\
\(T\) & Thrust magnitude & N & – \\
\(\mathbf{F}_{\text{prop}}^B \in \mathbb{R}^3\) & Propulsion force & N & \(B\) \\
\(\boldsymbol{\tau}_{\text{prop}}^B \in \mathbb{R}^3\) & Propulsion moment & N·m & \(B\) \\
\(\mathbf{F}_{\text{aero}}^B \in \mathbb{R}^3\) & Aerodynamic force & N & \(B\) \\
\(\mathbf{F}_{\text{aero}}^W \in \mathbb{R}^3\) & Aerodynamic force & N & \(W\) \\
\(C_D,C_L,C_Y\) & Force coefficients & – & – \\
\(C_\ell,C_m,C_n\) & Moment coefficients & – & – \\
\(\boldsymbol{\tau}_{\text{aero}}^B \in \mathbb{R}^3\) & Aero moment (net) & N·m & \(B\) \\
\(D_\omega\) & Damping matrix & N·m·s & \(B\) \\
\(\mathbf{f}_{\text{ext}}^W \in \mathbb{R}^3\) & External force (CoM) & N & \(W\) \\
\(\boldsymbol{\tau}_{\text{ext}}^B \in \mathbb{R}^3\) & External moment (CoM) & N·m & \(B\) \\
\(r_i^B \in \mathbb{R}^3\) & Surface pos. from CoM & m & \(B\) \\
\(\mathbf{v}_{a,i}^B \in \mathbb{R}^3\) & Local air velocity & m/s & \(B\) \\
\([\cdot]_\times\) & Skew-symmetric op. & – & – \\
\(\norm{\cdot}\) & Euclidean norm & – & – \\
\hline
\end{tabular}
\end{minipage}
\end{table*}
\endgroup

\begin{table*}[tb!]
\centering
\small
\caption{Model Summary: Kinematics, Aerodynamics, and Dynamics}
\label{tab:model_summary}
\renewcommand{\arraystretch}{2.1} 
\begin{tabularx}{\textwidth}{@{} >{\bfseries}l X @{}}
\toprule
Kinematics & 
$\dot{\mathbf{p}}^W = \mathbf{v}^W, \quad \dot{R} = R\,[\boldsymbol{\omega}^B]_\times$ \\

\hline

Relative wind & 
$\mathbf{v}_a^W = \mathbf{v}^W - \mathbf{w}^W, \quad \mathbf{v}_a^B = R^\top \mathbf{v}_a^W, \quad \mathbf{e}_a^B = \frac{\mathbf{v}_a^B}{\|\mathbf{v}_a^B\|}, \quad q = \tfrac{1}{2}\rho \|\mathbf{v}_a^B\|^2$ \newline
\textit{Regularization:} $\|\mathbf{v}_a^B\|_\epsilon = \max(\|\mathbf{v}_a^B\|, \epsilon), \quad \tilde{\mathbf{e}}_a^B = \frac{\mathbf{v}_a^B}{\|\mathbf{v}_a^B\|_\epsilon}, \quad \tilde{q} = \tfrac{1}{2}\rho\|\mathbf{v}_a^B\|_\epsilon^2$ \\

\hline

Aerodynamic directions (Geometric) & 
$\mathbf{e}_D^B = -\,\mathbf{e}_a^B, \quad \mathbf{e}_L^B = \frac{\mathbf{n}^B - (\mathbf{n}^B\cdot\mathbf{e}_a^B)\,\mathbf{e}_a^B}{\|\mathbf{n}^B - (\mathbf{n}^B\cdot\mathbf{e}_a^B)\,\mathbf{e}_a^B\|}, \quad \mathbf{e}_Y^B = \mathbf{e}_a^B \times \mathbf{e}_L^B$ \\

\hline

Aerodynamic Angles & 
$\alpha = \atanTwo\left(-\,\mathbf{e}_a^B\cdot\mathbf{n}^B,\ \mathbf{e}_a^B\cdot\mathbf{c}^B\right), \quad \beta = \atanTwo\left(-\mathbf{e}_a^B\cdot\mathbf{s}^B,\ \sqrt{1-(\mathbf{e}_a^B\cdot\mathbf{s}^B)^2}\right)$ \\

\hline

Aero Loads & 
$\mathbf{F}_{\text{aero}}^B = q\,S\left( C_D\,\mathbf{e}_D^B + C_L\,\mathbf{e}_L^B + C_Y\,\mathbf{e}_Y^B \right), \quad \mathbf{F}_{\text{aero}}^W = R\,\mathbf{F}_{\text{aero}}^B$ \newline
$\boldsymbol{\tau}_{\text{aero}}^B = q\,S\left( C_\ell\,b\,\mathbf{c}^B + C_m\,\bar c\,\mathbf{s}^B + C_n\,b\,\mathbf{n}^B \right) + D_\omega\,\boldsymbol{\omega}^B$ \\

\hline

Propulsion \& External & 
$\mathbf{F}_{\text{prop}}^B = T\,\mathbf{u}_t^B, \quad \boldsymbol{\tau}_{\text{prop}}^B = \boldsymbol{\tau}_{\text{prop}}^B(\mathbf{u})$ \newline
$\mathbf{f}_{\text{ext}}^W \text{ (Force at } r^B\text{)}, \quad \boldsymbol{\tau}_{\text{ext,net}}^B = \boldsymbol{\tau}_{\text{ext}}^B + r^B \times (R^\top \mathbf{f}_{\text{ext}}^W)$ \\

\hline

Rigid-Body Dynamics & 
$m\,\dot{\mathbf{v}}^W = m\,\mathbf{g}^W + R(\mathbf{F}_{\text{prop}}^B + \mathbf{F}_{\text{aero}}^B) + \mathbf{f}_{\text{ext}}^W$ \newline
$I_B\,\dot{\boldsymbol{\omega}}^B + \boldsymbol{\omega}^B \times (I_B \boldsymbol{\omega}^B) = \boldsymbol{\tau}_{\text{prop}}^B + \boldsymbol{\tau}_{\text{aero}}^B + \boldsymbol{\tau}_{\text{ext,net}}^B$ \\

\hline

Special Cases & 
\textbf{Distributed surfaces:} $\mathbf{v}_{a,i}^B = \mathbf{v}_a^B + \boldsymbol{\omega}^B \times r_i^B, \quad \boldsymbol{\tau}_{\text{aero}}^B = \sum_i r_i^B \times \mathbf{F}_i^B$ \newline
\textbf{Coordinated flight:} Enforce $\beta=0$ ($\mathbf{s}^W \perp \mathbf{v}^W$) and solve planar balance for $T, \alpha$ \\
\bottomrule
\end{tabularx}
\end{table*}

\section{Aerodynamic Coordination vs. Perceived Coordination in  Tethered Flight}

In standard free flight, a “coordinated turn” implies two simultaneous conditions: zero aerodynamic sideslip (\(\beta=0\)) and zero lateral acceleration felt by the pilot (the “ball” is centered). In tethered flight, these two conditions decouple.
Our derivation enforces \(\beta=0\) for aerodynamic efficiency. However, a pilot or accelerometer on board measures the \emph{specific force} \(\mathbf{f}_{\text{meas}}^B = R(t)^\top\big(\mathbf{a}^W - \mathbf{g}^W\big)\), which is determined purely by the kinematics.
In contrast, the aircraft’s attitude is aligned with the required net force \(\mathbf{F}_{\text{req}}^W\), which implies that the lift vector opposes the sum of gravity, inertial forces, \emph{and} the tether force. Because \(\mathbf{F}_{\text{ext}} \neq \mathbf{0}\), the aircraft’s body \(z\)-axis does not align with the specific force vector \(\mathbf{f}_{\text{meas}}^B\).
Consequently, while the aircraft flies cleanly through the air, a pilot would feel a distinct lateral force pushing them sideways in their seat—the “ball” would not be centered, reflecting the tension of the tether rather than an uncoordinated aerodynamic slide.

\subsubsection*{(Optional) Discussion: Practical Implications of the Decoupled Equilibrium.}

Maintaining aerodynamic coordination (\(\beta=0\)) is of paramount importance for the aircraft’s performance and safety. It ensures “clean airflow” along the fuselage, minimizing separation drag, preventing asymmetric stall risks, ensuring symmetric intake airflow for propulsion efficiency, and avoiding structural fatigue on the vertical stabilizer. However, achieving this aerodynamic purity while tethered creates a decoupling between the airframe attitude and the forces felt onboard, leading to several engineering challenges:

\begin{itemize}
    \item \emph{Fluid Systems:} Conventional fuel tanks rely on the specific force vector aligning with the tank bottom (coordinated flight) to keep fuel at the pump intakes. The persistent lateral acceleration in tethered flight requires pressurized bladders or aerobatic header tanks to prevent fuel starvation (unporting).
    \item \emph{State Estimation:} Standard Attitude and Heading Reference Systems (AHRS) often assume that the specific force vector aligns with the body vertical axis during coordinated maneuvers to correct gyroscope drift. The inherent lateral specific force in this trajectory will confuse standard Extended Kalman Filters (EKF), necessitating an estimator that explicitly models the external tether force.
    \item \emph{Ergonomics:} A pilot or sensitive payload will experience a constant lateral load. This can be mitigated by banking the seat or payload mount relative to the airframe by the difference between the aerodynamic bank angle and the apparent gravity vector, ensuring forces are loaded purely in the sagittal plane.
\end{itemize}

\section{Parameter Study and Assumptions for Representative UAV Classes}
\label{app:param-study}

\begin{table*}[t]
\centering
\caption{Recommended parameter ranges and nominal values for two fixed-wing UAV classes. See the rightmost column for the sources supporting each number.}
\label{tab:sim_params}
\renewcommand{\arraystretch}{1.15}
\begin{tabular}{l l l l l}
\toprule
\textbf{Parameter} & \textbf{Symbol} & \textbf{Class A (1–3 m)} & \textbf{Class B (4–6 m)} & \textbf{Sources} \\
\midrule
Span & $b$ [m] & $1$–$3$ (e.g., $2.12$) & $4$–$6$ (e.g., $5.00$) & \cite{SkywalkerX8,InsituScanEagle,TextronAerosonde,AirMobiSkyeye} \\
Reference area & $S$ [m$^2$] & $0.2$–$0.9$ (e.g., $0.80$) & $1.3$–$4.5$ (e.g., $2.615$) & \cite{SkywalkerX8,AirMobiSkyeye} \\
Aspect ratio & $\mathrm{AR}$ [-] & $\approx 5$–$7$ (e.g., $5.62$) & $\approx 8$–$12$ (e.g., $9.56$) & \cite{UIUC_AR,AirMobiSkyeye} \\
Oswald eff. & $e$ [-] & $0.75$–$0.85$ (nom.\ $0.80$) & $0.85$–$0.92$ (nom.\ $0.90$) & \cite{Anderson,MIT_lecture} \\
Lift slope & $C_{L\alpha}$ [rad$^{-1}$] & $4.2$–$4.7$ (nom.\ $4.35$) & $4.9$–$5.3$ (nom.\ $5.10$) & \cite{Anderson,MIT_lecture} \\
Zero-lift drag & $C_{D0}$ [-] & $0.03$–$0.05$ (nom.\ $0.035$) & $0.02$–$0.04$ (nom.\ $0.030$) & \cite{AeroToolboxPolar} \\
Induced factor & $k=1/(\pi e\,\mathrm{AR})$ & $0.05$–$0.08$ (nom.\ $0.0708$) & $0.03$–$0.05$ (nom.\ $0.0370$) & \cite{NASADrag,AeroToolboxPolar} \\
Quadratic in $\alpha$ & $k_\alpha=k\,C_{L\alpha}^2$ [rad$^{-2}$] & $1.1$–$1.4$ (nom.\ $1.34$) & $0.83$–$1.10$ (nom.\ $0.962$) & \cite{Anderson,MIT_lecture} \\
Mass (typical) & $m$ [kg] & $0.7$–$3.5$ & $15$–$95$ & \cite{eBeeX,InsituScanEagle,TextronAerosonde,SilentFalcon,AirMobiSkyeye,Mugin5000} \\
Cruise speed & $V$ [m/s] & $15$–$25$ (nom.\ $18$) & $25$–$30$ (nom.\ $27$) & \cite{eBeeX,InsituScanEagle,TextronAerosonde} \\
Dyn.\ pressure & $q=\tfrac12 \rho V^2$ [Pa] & $138$–$383$ (nom.\ $198$) & $383$–$551$ (nom.\ $447$) & \cite{ISA_Cambridge} \\
\bottomrule
\end{tabular}
\end{table*}

This appendix documents the procedure by which we obtained the “realistic” aerodynamic and sizing parameters summarized in Table~\ref{tab:sim_params} for two classes of fixed-wing UAVs—Class~A with wingspan $b\!=\!1\text{--}3$~m and Class~B with $b\!=\!4\text{--}6$~m—and how these parameters tie into the geometric Newton--Euler model in Section~\ref{sec:model} and the inverse-dynamics construction in Section~\ref{sec:inverse}.

\subsection*{Scope and linkage to the main text}
The parameters in Table~\ref{tab:sim_params} are the quantities that directly enter:
(i) the dynamic pressure and force model in \eqref{eq:q} and \eqref{eq:FaeroW}, via the pair $(q,S)$ and the lift--drag maps $C_L(\alpha)$, $C_D(\alpha)$; and
(ii) the inverse simulation algebra in \eqref{eq:Faero_coord_wind}--\eqref{eq:T_alpha_closed_wind} and \eqref{eq:alpha_scalar_eq}, where $C_{L\alpha}$, $C_{D0}$ and the quadratic coefficient in $\alpha$ determine the 2D force balance.
All \emph{forces} are composed in the world frame and \emph{moments} in the body frame as per Sections~\ref{sec:model}--\ref{sec:inverse}. We retain the FLU convention discussed in the Convention note; when cross-referencing FRD tables, the usual $y$-axis sign considerations apply (cf.\ footnote below \eqref{eq:moment_coeffs_wind}).

\subsection*{Representative platforms and geometric baselines}

To anchor the two span classes to real aircraft, we surveyed well-documented platforms:
\begin{itemize}
    \item for Class~A, the senseFly eBee family ($b\!\approx\!1.1$--$1.2$~m, cruise $V\!\approx\!11$--$30$~m/s) and the Skywalker X8 flying wing ($b\!=\!2.12$~m, published $S\!=\!0.80$~m$^2$) \cite{eBeeX,SkywalkerX8};
\item for Class~B, ScanEagle ($b\!\approx\!3.1$~m, endurance 18--20~h), Silent Falcon ($b\!\approx\!4.4$~m), and Aerosonde Mk.~4.7 ($b\!\approx\!4.4$~m, endurance 12--18~h), complemented by the 5~m ``Skyeye/Mugin'' frames which publish wing area around $S\!\approx\!2.6$~m$^2$ \cite{InsituScanEagle,SilentFalcon,TextronAerosonde,AirMobiSkyeye,Mugin5000}.
\end{itemize}
From these, we formed ranges for span $b$, reference area $S$, typical mass $m$, and cruise speed $V$ used to compute $q$ via \eqref{eq:q}.

\subsection*{Aerodynamic model and small-angle approximations}

Throughout Section~\ref{sec:inverse} we adopt the standard small-angle polar\footnote{In aerodynamics, a \emph{polar} is the relationship between nondimensional force coefficients—most commonly the drag polar, which relates $C_D$ to $C_L$ (e.g., $C_D=C_{D0}+k\,C_L^2$) or, equivalently, to the operating state such as the angle of attack $\alpha$. It is not a full dynamic model; rather, it plays the role of a \emph{constitutive law} that supplies lift/drag coefficients to the Newton–Euler equations in Section~\ref{sec:model}. By \emph{small-angle polar} we mean the pre-stall, low-Mach approximation about $\alpha=0$: $C_L\!\approx\!C_{L\alpha}\alpha$ and $C_D\!\approx\!C_{D0}+k_\alpha\alpha^2$ with $k_\alpha=k\,C_{L\alpha}^2$ (the classical parabolic drag polar) \cite{Anderson,NASADrag,AeroToolboxPolar}.}
\[
C_L(\alpha)\approx C_{L\alpha}\,\alpha,\qquad
C_D(\alpha)\approx C_{D0} + k_\alpha\,\alpha^2,
\]
valid in the pre-stall regime (typically up to $\alpha\sim8^\circ$--$12^\circ$, airfoil/Re dependent) \cite{MIT_lecture,Anderson}.
The parameters $C_{L\alpha}$, $C_{D0}$ and $k_\alpha$ are not arbitrary: they are derived from finite-wing theory and the parabolic drag polar, as detailed next.

\subsection*{Computation of $C_{L\alpha}$, $k$, and $k_\alpha$}
For a finite unswept wing, we use the textbook relation for lift-curve slope
\begin{equation}
  C_{L\alpha}\;=\;\frac{a_0}{1+\dfrac{a_0}{\pi e\,\mathrm{AR}}},\qquad a_0\approx 2\pi~\text{rad}^{-1},
  \label{eq:CLalpha_finite}
\end{equation}
where $\mathrm{AR}=b^2/S$ is the aspect ratio and $e\in(0,1]$ is the Oswald efficiency factor \cite{MIT_lecture,Anderson,TUdelft_LiftSlope}. This yields the Class~A range $C_{L\alpha}\!\approx\!4.2$--$4.7~\text{rad}^{-1}$ for $\mathrm{AR}\!\approx\!5$--$7$ and $e\!\approx\!0.75$--$0.85$, and the Class~B range $C_{L\alpha}\!\approx\!4.9$--$5.3~\text{rad}^{-1}$ for $\mathrm{AR}\!\approx\!8$--$12$ and $e\!\approx\!0.85$--$0.92$ (see Table~\ref{tab:sim_params}). These bands are consistent with low-to-moderate AR UAV literature and wind-tunnel data at low $Re$ \cite{UIUC_AR,MIT_lecture}.

For drag, we adopt the parabolic polar
\begin{equation}
  C_D \;=\; C_{D0} + k\,C_L^2,\qquad k=\frac{1}{\pi e\,\mathrm{AR}},
  \label{eq:parabolic}
\end{equation}
which is appropriate for the speed/Mach numbers of interest \cite{NASADrag,AeroToolboxPolar}. Substituting $C_L\!\approx\!C_{L\alpha}\alpha$ gives the coefficient $k_\alpha$ used in Section~\ref{sec:inverse}:
\begin{equation}
  k_\alpha \;=\; k\,C_{L\alpha}^2 \;=\; \frac{C_{L\alpha}^2}{\pi e\,\mathrm{AR}}.
  \label{eq:kalphadef}
\end{equation}
Numerically, this produces $k_\alpha\!\approx\!1.1$--$1.4~\text{rad}^{-2}$ (Class~A) and $k_\alpha\!\approx\!0.83$--$1.10~\text{rad}^{-2}$ (Class~B); see Table~\ref{tab:sim_params}.

\subsection*{Choice of $C_{D0}$ and dynamic pressure $q$}

The zero-lift drag $C_{D0}$ aggregates profile, form, interference and excrescence contributions. Following standard preliminary-estimation practice, we selected the ranges
$C_{D0}\!\in\![0.03,0.05]$ (Class~A foam/composite small UAVs at low $Re$) and
$C_{D0}\!\in\![0.02,0.04]$ (cleaner endurance/tactical airframes),
consistent with drag build-ups and historical data \cite{AeroToolboxPolar}.
For $q$ we use the ISA sea-level density $\rho=1.225~\mathrm{kg/m^3}$ and the published cruise-speed bands of the representative aircraft to compute $q=\frac12\rho V^2$:
Class~A $V\!\in\![15,25]$~m/s $\Rightarrow q\!\in\![138,383]$~Pa;
Class~B $V\!\in\![25,30]$~m/s $\Rightarrow q\!\in\![383,551]$~Pa \cite{ISA_Cambridge,eBeeX,InsituScanEagle,TextronAerosonde}.

\subsection*{Nominal values used in figures (reproducibility)}

For definiteness in the examples of this work, we instantiate one nominal point per class (centered within the ranges above):
\begin{itemize}
  \item \textbf{Class A (Skywalker X8-like).} $(b,S,m,V)=(2.12~\mathrm{m},\,0.80~\mathrm{m^2},\,3.0~\mathrm{kg},\,18~\mathrm{m/s})$,
        $\mathrm{AR}=5.62$, $e=0.80$,
        $C_{L\alpha}=4.35~\mathrm{rad^{-1}}$,
        $C_{D0}=0.035$,
        $k=\frac{1}{\pi e\,\mathrm{AR}}=0.0708$,
        $k_\alpha=1.34~\mathrm{rad^{-2}}$,
        $q=198~\mathrm{Pa}$ \cite{SkywalkerX8,MIT_lecture,AeroToolboxPolar,ISA_Cambridge}.
  \item \textbf{Class B (Skyeye 5 m-like).} $(b,S,m,V)=(5.00~\mathrm{m},\,2.615~\mathrm{m^2},\,40~\mathrm{kg},\,27~\mathrm{m/s})$,
        $\mathrm{AR}=9.56$, $e=0.90$,
        $C_{L\alpha}=5.10~\mathrm{rad^{-1}}$,
        $C_{D0}=0.030$,
        $k=0.0370$,
        $k_\alpha=0.962~\mathrm{rad^{-2}}$,
        $q=447~\mathrm{Pa}$ \cite{AirMobiSkyeye,MIT_lecture,AeroToolboxPolar,ISA_Cambridge}.
\end{itemize}
These nominal sets are used to plot the polar and to solve \eqref{eq:two_eqs_wind}--\eqref{eq:T_alpha_closed_wind} and \eqref{eq:alpha_scalar_eq} in closed form.

\subsection*{Validity range, sensitivity, and iteration with control effects}

The linear $C_L(\alpha)$ and parabolic $C_D(C_L)$ models are valid in the pre-stall region and at low Mach numbers. For the platforms and speeds considered here, compressibility is negligible and the linear slope tracks thin-airfoil theory closely before stall \cite{MIT_lecture,Anderson}.
As noted in Step~5 of Section~\ref{sec:inverse}, we decouple control-surface effects from the translational force model; if large flap/elevon deflections are used (e.g., high-lift configurations), we recommend iterating Steps~5--9 and updating $C_L(\alpha,\mathbf{u})$, $C_D(\alpha,\mathbf{u})$ accordingly.
Sensitivity to $e$ and $\mathrm{AR}$ appears via \eqref{eq:CLalpha_finite} and \eqref{eq:kalphadef}; Table~\ref{tab:sim_params} brackets these effects.

\subsection*{Units, frames, and sign conventions}

All coefficients are nondimensional and defined with the reference area $S$ and the dynamic pressure $q$ as in \eqref{eq:FaeroB}--\eqref{eq:FaeroW}.
Coefficients and directions in Section~\ref{sec:model} are expressed in the robotics FLU convention; when comparing to FRD-based aeronautics tables, the $y$-axis sign flip must be applied for pitch/yaw moments (see footnote below \eqref{eq:moment_coeffs_wind}).

\medskip
\noindent\textbf{Data provenance.}
Platform geometries ($b,S$), masses $m$, and cruise speeds $V$ are taken from manufacturer data sheets and widely cited summaries for eBee, Skywalker X8, ScanEagle, Silent Falcon, Aerosonde Mk.~4.7, and Skyeye/Mugin 5~m frames \cite{eBeeX,SkywalkerX8,InsituScanEagle,SilentFalcon,TextronAerosonde,AirMobiSkyeye,Mugin5000}. Aerodynamic relations for $C_{L\alpha}$ and the parabolic polar are standard \cite{MIT_lecture,Anderson,NASADrag,AeroToolboxPolar}, and the ISA density used in \eqref{eq:q} is given in \cite{ISA_Cambridge}.

\printbibliography[title={Appendix References}]

\end{refsection}

\fi

\end{document}